%% file: main.tex
\documentclass{article}

\usepackage{subcaption}
\usepackage[final,nonatbib]{neurips_data_2021}

\usepackage[utf8]{inputenc}
\usepackage[T1]{fontenc}
\usepackage{hyperref}
\usepackage{url}
\usepackage{booktabs}
\usepackage{amsfonts}
\usepackage{nicefrac}
\usepackage{microtype}
\usepackage{xcolor}
\usepackage{multirow}
\usepackage{adjustbox}
\usepackage{wrapfig}
\usepackage{tablefootnote}
\usepackage{enumitem}
\usepackage{authblk}
\usepackage{siunitx}
\usepackage{float}
\usepackage{cleveref}

\newcommand{\sectioncolor}{violet}
\usepackage{mdframed}

\makeatletter
\renewcommand\AB@affilsepx{, \protect\Affilfont}
\makeatother

\Crefname{figure}{\textbf{Figure}}{}

\title{Argoverse 2: Next Generation Datasets for Self-Driving Perception and Forecasting}

\author{
  \textbf{Benjamin Wilson}$^{* \dag, 1}$, \hspace{1mm} \textbf{William Qi}$^{* \dag}$, \hspace{1mm} \textbf{Tanmay Agarwal}$^{* \dag}$, \hspace{1mm} \textbf{John Lambert}$^\dag$, \hspace{1mm} \textbf{Jagjeet Singh}$^\dag$, \\ 
  \textbf{Siddhesh Khandelwal}$^2$, \hspace{1mm} \textbf{Bowen Pan}$^{\dag, 3}$, \hspace{1mm} \textbf{Ratnesh Kumar}$^\dag$, \hspace{1mm} \textbf{Andrew Hartnett}$^\dag$, \\ \vspace{3.5mm}
  \textbf{Jhony Kaesemodel Pontes}$^\dag$, \hspace{1mm} \textbf{Deva Ramanan}$^{\dag, 4}$, \hspace{1mm} \textbf{Peter Carr}$^\dag$, \hspace{1mm} \textbf{James Hays}$^{\dag,1}$\hspace{1mm} \\
  \vspace{3mm} $^1$Georgia Tech,\hspace{1mm} $^2$UBC,\hspace{1mm} $^3$MIT,\hspace{1mm} $^4$CMU\hspace{1mm}
}

\begin{document}

\maketitle

\def\thefootnote{}\footnotetext{*Equal contribution.}\def\thefootnote{\arabic{footnote}}

\def\thefootnote{}\footnotetext{$^\dag$Work completed while at Argo AI.}\def\thefootnote{\arabic{footnote}}

\begin{abstract}
 We introduce Argoverse 2 (AV2) --- a collection of three datasets for perception and forecasting research in the self-driving domain. The annotated \emph{Sensor Dataset} contains 1,000 sequences of multimodal data, encompassing high-resolution imagery from seven ring cameras, and two stereo cameras in addition to lidar point clouds, and 6-DOF map-aligned pose. Sequences contain 3D cuboid annotations for 26 object categories, all of which are sufficiently-sampled to support training and evaluation of 3D perception models. The \emph{Lidar Dataset} contains 20,000 sequences of unlabeled lidar point clouds and map-aligned pose. This dataset is the largest ever collection of lidar sensor data and supports self-supervised learning and the emerging task of point cloud forecasting. Finally, the \emph{Motion Forecasting Dataset} contains 250,000 scenarios mined for interesting and challenging interactions between the autonomous vehicle and other actors in each local scene. Models are tasked with the prediction of future motion for ``scored actors" in each scenario and are provided with track histories that capture object location, heading, velocity, and category. In all three datasets, each scenario contains its own \emph{HD Map} with 3D lane and crosswalk geometry --- sourced from data captured in six distinct cities. We believe these datasets will support new and existing machine learning research problems in ways that existing datasets do not. All datasets are released under the CC BY-NC-SA 4.0 license.
\end{abstract}

\section{Introduction}

\input{neurips_data_2021/sections/introduction}
\input{neurips_data_2021/sections/related-work}

\input{neurips_data_2021/sections/dataset}
\input{neurips_data_2021/sections/evaluation}
\input{neurips_data_2021/sections/conclusion}

\bibliographystyle{plain}
\bibliography{references.bib}

\input{neurips_data_2021/sections/supplementary.tex}

\input{neurips_data_2021/sections/datasheet.tex}

\end{document}

%% file: neurips_data_2021/sections/introduction.tex
In order to achieve the goal of safe, reliable autonomous driving, a litany of machine learning tasks must be addressed, from stereo depth estimation to motion forecasting to 3D object detection. In recent years, numerous high quality self-driving datasets have been released to support research into these and other important machine learning tasks. Many datasets are annotated ``sensor'' datasets~\cite{Caesar20cvpr_NuScenes,Sun20cvpr_WaymoOpenDataset,H3D,A3D,lyft_sensor,ApolloScape,a2d2,Cityscapes3D,Pitropov_2020,Mao21arxiv_ONCEOneMillionScenes} in the spirit of the influential KITTI dataset~\cite{Geiger12cvpr_KITTIAreWeReady}. The Argoverse 3D Tracking dataset~\cite{Chang19cvpr_Argoverse} was the first such dataset with ``HD maps'' --- maps containing lane-level geometry. Also influential are self-driving ``motion prediction'' datasets~\cite{Ettinger21arxiv_WaymoOpenMotion,Houston20arxiv_LyftL5,Malinin21arxiv_ShiftsYandex,Caesar20cvpr_NuScenes,zhan2019interaction} --- containing abstracted object tracks instead of raw sensor data --- of which the Argoverse Motion Forecasting dataset~\cite{Chang19cvpr_Argoverse} was the first.

In the last two years, the Argoverse team has hosted six competitions on 3D tracking, stereo depth estimation, and motion forecasting. We maintain evaluation servers and leaderboards for these tasks, as well as 3D detection. The leaderboards collectively contain thousands of submissions from four hundred teams\footnote{This count includes private submissions not posted to the public leaderboards.}. We also maintain the Argoverse API and have addressed more than one hundred issues\footnote{\url{https://github.com/argoverse/argoverse-api}}. From these experiences we have formed the following guiding principles to guide the creation of the next iteration of Argoverse datasets.

\begin{enumerate}[leftmargin=*]
    \item \textbf{Bigger isn't always better.} Self-driving vehicles capture a flood of sensor data which is logistically difficult to work with. Sensor datasets are several terabytes in size, even when compressed. If standard benchmarks grow further, we risk alienating much of the academic community and leaving progress to well-resourced industry groups. \emph{For this reason, we match but do not exceed the scale of sensor data in nuScenes~\cite{Caesar20cvpr_NuScenes} and Waymo Open~\cite{Sun20cvpr_WaymoOpenDataset}}.
    \item \textbf{Make every instance count.} Much of driving is boring. Datasets should focus on the difficult, interesting scenarios where current forecasting and perception systems struggle. \emph{Therefore we mine for especially crowded, dynamic, and kinematically unusual scenarios.}
    \item \textbf{Diversity matters.} Training on data from wintertime Detroit is not sufficient for detecting objects in Miami --- Miami has 15 times the frequency of motorcycles and mopeds. Behaviors differ as well, so learned pedestrian motion behavior might not generalize. \emph{Accordingly, each of our datasets are drawn from six diverse cities --- Austin, Detroit, Miami, Palo Alto, Pittsburgh, and Washington D.C. --- and different seasons, as well, from snowy to sunny}.
    \item \textbf{Map the world.} HD maps are  powerful priors for perception and forecasting. Learning-based methods that found clever ways to encode map information~\cite{Liang20eccv_LaneGCN} performed well in Argoverse competitions. \emph{For this reason, we augment our HD map representation with 3D lane geometry, paint markings, crosswalks, higher resolution ground height, and more}.
    \item \textbf{Self-supervise.} Other machine learning domains have seen enormous success from self-supervised learning in recent years. Large-scale lidar data from dynamic scenes, paired with HD maps, could lead to better representations than current supervised approaches. \emph{For this reason, we build the largest dataset of lidar sensor data.}
    \item \textbf{Fight the heavy tail.} Passenger vehicles are common, and thus we can assess our forecasting and detection accuracy for cars. However, with existing datasets, we cannot assess forecasting accuracy for buses and motorcycles with their distinct behaviors, nor can we evaluate stroller and wheel chair detection. \emph{Thus we introduce the largest taxonomy to date for sensor and forecasting datasets, and we ensure enough samples of rare objects to train and evaluate models.}
\end{enumerate}

With these guidelines in mind we built the three Argoverse 2 (AV2) datasets. Below, we highlight some of their contributions.

\begin{enumerate}[leftmargin=*]
    \item The 1,000 scenario \emph{Sensor dataset} has the largest self-driving taxonomy to date -- 30 categories. 26  categories contain at least 6,000 cuboids to enable diverse taxonomy training and testing. The dataset also has stereo imagery, unlike recent self-driving datasets.
    \item The 20,000 scenario \emph{Lidar dataset} is the largest dataset for self-supervised learning on lidar. The only similar dataset, concurrently developed ONCE~\cite{Mao21arxiv_ONCEOneMillionScenes}, does not have HD maps.
    \item The 250,000 scenario \emph{Motion Forecasting Dataset} has the largest taxonomy -- 5 types of dynamic actors and 5 types of static actors -- and covers the largest mapped area of any such dataset.
\end{enumerate}

We believe these datasets will support research into problems such as 3D detection, 3D tracking, monocular and stereo depth estimation, motion forecasting, visual odometry, pose estimation, lane detection, map automation, self-supervised learning, structure from motion, scene flow, optical flow, time to contact estimation, and point cloud forecasting. 

%% file: neurips_data_2021/sections/related-work.tex
\section{Related Work}
\vspace{-4mm}

The last few years have seen rapid progress in self-driving perception and forecasting research, catalyzed by many high quality datasets.

\textbf{Sensor datasets and 3D Object Detection and Tracking.} New sensor datasets for 3D object detection~\cite{Caesar20cvpr_NuScenes,Sun20cvpr_WaymoOpenDataset,H3D,A3D,lyft_sensor,ApolloScape,a2d2,Cityscapes3D,Pitropov_2020,Mao21arxiv_ONCEOneMillionScenes} have led to influential detection methods such as  anchor-based approaches like PointPillars \cite{Lang19cvpr_PointPillars}, and more recent anchor-free approaches such as AFDet \cite{Ge20cvprw_AFDet} and CenterPoint \cite{Yin21cvpr_CenterPoint}. These methods have led to dramatic accuracy improvements on all datasets. In turn, these improvements have made isolation of object-specific point clouds possible, which has proven invaluable for  offboard detection and tracking \cite{Qi21cvpr_Offboard3dObjectDetection}, and for simulation \cite{Chen21cvpr_GeoSim}, which previously required human-annotated 3D bounding boxes \cite{Manivasagam20cvpr_LiDARsim}. New approaches explore alternate point cloud representations, such as range images \cite{Chai21cvpr_RangeImageGraphConvolution,Bewley20corl_RangeDilatedConvolutions,Sun21cvpr_RangeSparseNet}. Streaming perception \cite{Li20eccv_StreamingPerception,Han20eccv_StreamingDetectionPointClouds} introduces a paradigm to explore the tradeoff between accuracy and latency. A detailed comparison between the AV2 \emph{Sensor Dataset} and recent 3D object detection datasets is provided in Table~\ref{tab:sensor-datasets-comparison}.

\textbf{Motion Forecasting.} 
For motion forecasting, the progress has been just as significant. A transition to attention-based methods \cite{Leurent19arxiv_SocialAttention, Mercat19arxiv_MultiheadAttentionForecasting, Mercat20icra_MultiheadAttentionForecasting} has led to a variety of new vector-based representations for map and trajectory data \cite{Gao20cvpr_VectorNet,Liang20eccv_LaneGCN}. New datasets have also paved the way for new algorithms, with nuScenes \cite{Caesar20cvpr_NuScenes}, Lyft L5 \cite{Houston20arxiv_LyftL5}, and the Waymo Open Motion Dataset \cite{Ettinger21arxiv_WaymoOpenMotion} all releasing lane graphs after they proved to be essential in Argoverse 1~\cite{Chang19cvpr_Argoverse}. Lyft  also introduced traffic/speed control data, while Waymo added crosswalk polygons, lane boundaries (with marking type), speed limits, and stop signs to the map. More recently, Yandex has released the Shifts \cite{Malinin21arxiv_ShiftsYandex} dataset, which is the largest (by scenario hours) collection of forecasting data available to date. Together, these datasets have enabled exploration of multi-actor, long-range motion forecasting leveraging both static and dynamic maps.

Following upon the success of Argoverse 1.1, we position AV2 as a large-scale repository of high-quality motion forecasting scenarios - with guarantees on data frequency (exactly 10 Hz) and diversity (>2000 km of unique roadways covered across 6 cities). This is in contrast to nuScenes (reports data at just 2 Hz) and Lyft (collected on a single 10 km segment of road), but is complementary to Waymo Open Motion Dataset (employs a similar approach for scenario mining and data configuration). Complementary datasets are essential for these safety critical problems as they provide opportunities to evaluate generalization and explore transfer learning. To improve ease of use, we have also designed AV2 to be widely accessible both in terms of data size and format --- a detailed comparison vs. other recent forecasting datasets is provided in Table \ref{forecasting-comparison}.

\textbf{Broader Problems of Perception for Self-Driving.}
Aside from the tasks of object detection and motion forecasting, new, large-scale sensor datasets for self-driving present opportunities to explore dozens of new problems for perception, especially those that can be potentially solved via self-supervision.  A number of new problems have been recently proposed; real-time 3D semantic segmentation in video has received attention thanks to SemanticKITTI \cite{Behley19iccv_SemanticKITTI}. HD map automation \cite{Zurn21arxiv_LaneGraphEstimation,Li21arxiv_HDMapNet} and HD map change detection \cite{Lambert21neurips_TrustButVerifyHDMapChangeDetection} have received additional attention, along with 3D scene flow and pixel-level scene simulation \cite{Yang20cvpr_SurfelGAN,Chen21cvpr_GeoSim}. Datasets exist with unique modalities such as thermal imagery~\cite{ChoiThermal1,ChoiThermal2}. Our new \emph{Lidar Dataset} enables large-scale self-supervised training of new approaches for freespace forecasting \cite{Hu21cvpr_SelfSupervisedFreespaceForecasting} or point cloud forecasting \cite{Weng20eccvw_SPF, Weng20corl_SPF2SequentialPointcloudForecasting}.

%% file: neurips_data_2021/sections/dataset.tex
\section{The Argoverse 2 Datasets}

\subsection{Sensor Dataset}

\begin{figure}
    \centering
    \vspace{-3mm}
    \includegraphics[trim={0 0 0 1.5cm},clip,width=1.0\linewidth]{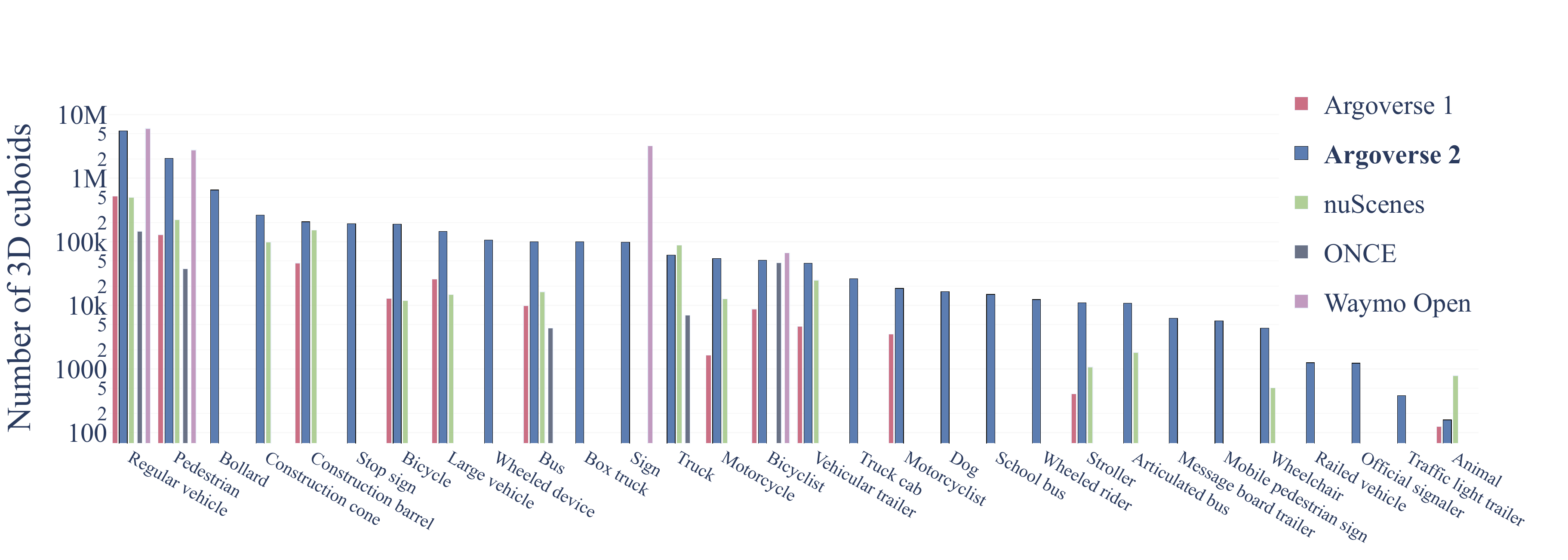}
    \vspace{-7mm}
    \caption{Number of annotated 3D cuboids per category for Argoverse 1 \emph{3D Tracking}, Argoverse 2 \emph{Sensor Dataset}, nuScenes, ONCE, and Waymo Open. The nuScenes annotation rate is \SI{2}{\hertz}, compared to \SI{10}{\hertz} for Argoverse, but that does not account for the relative increase in object diversity in Argoverse 2.}
    \label{fig:cuboids-stats}
\end{figure}

\begin{figure}
    \centering
    \includegraphics[width=0.45\linewidth]{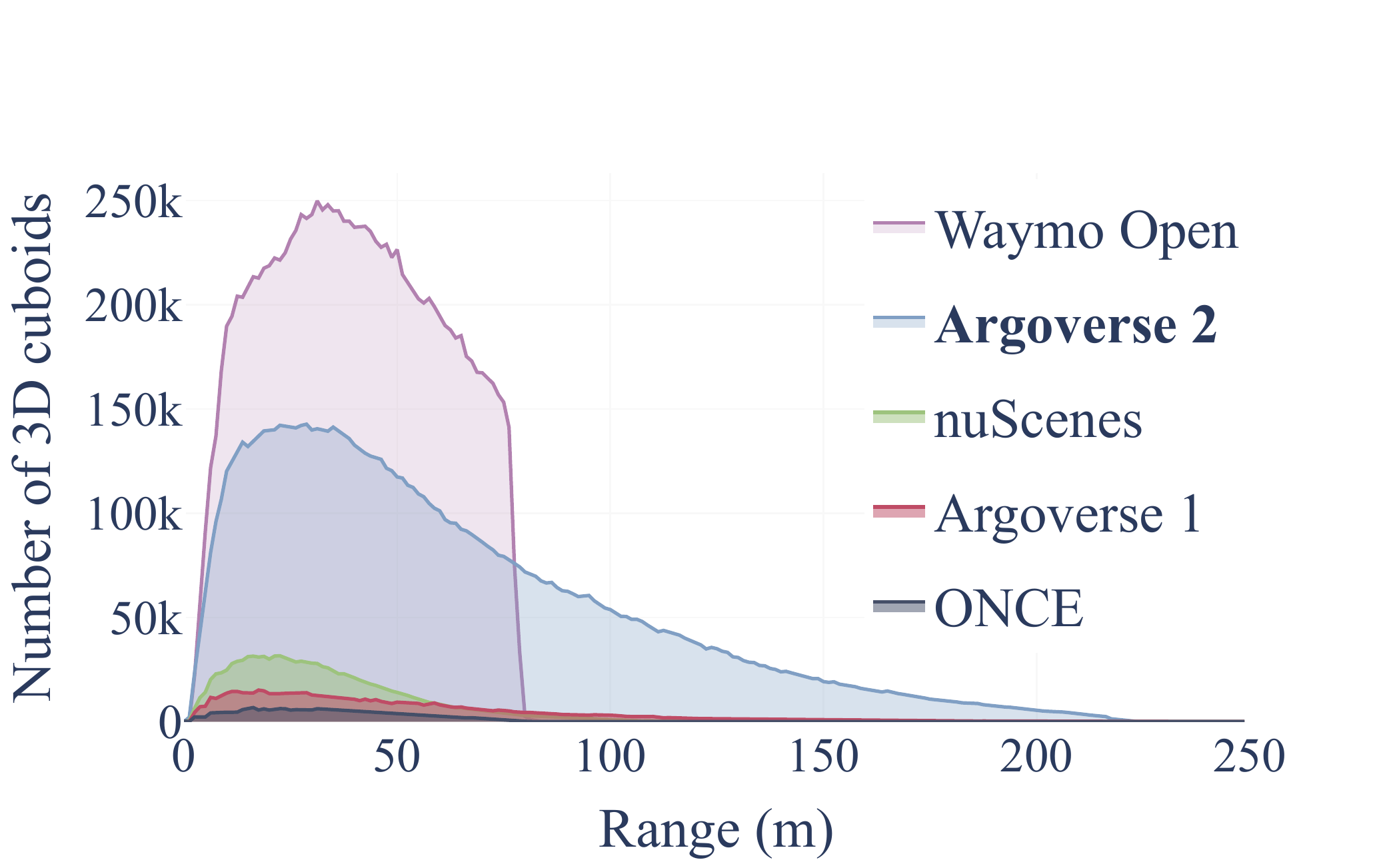}\qquad
    \includegraphics[width=0.45\linewidth]{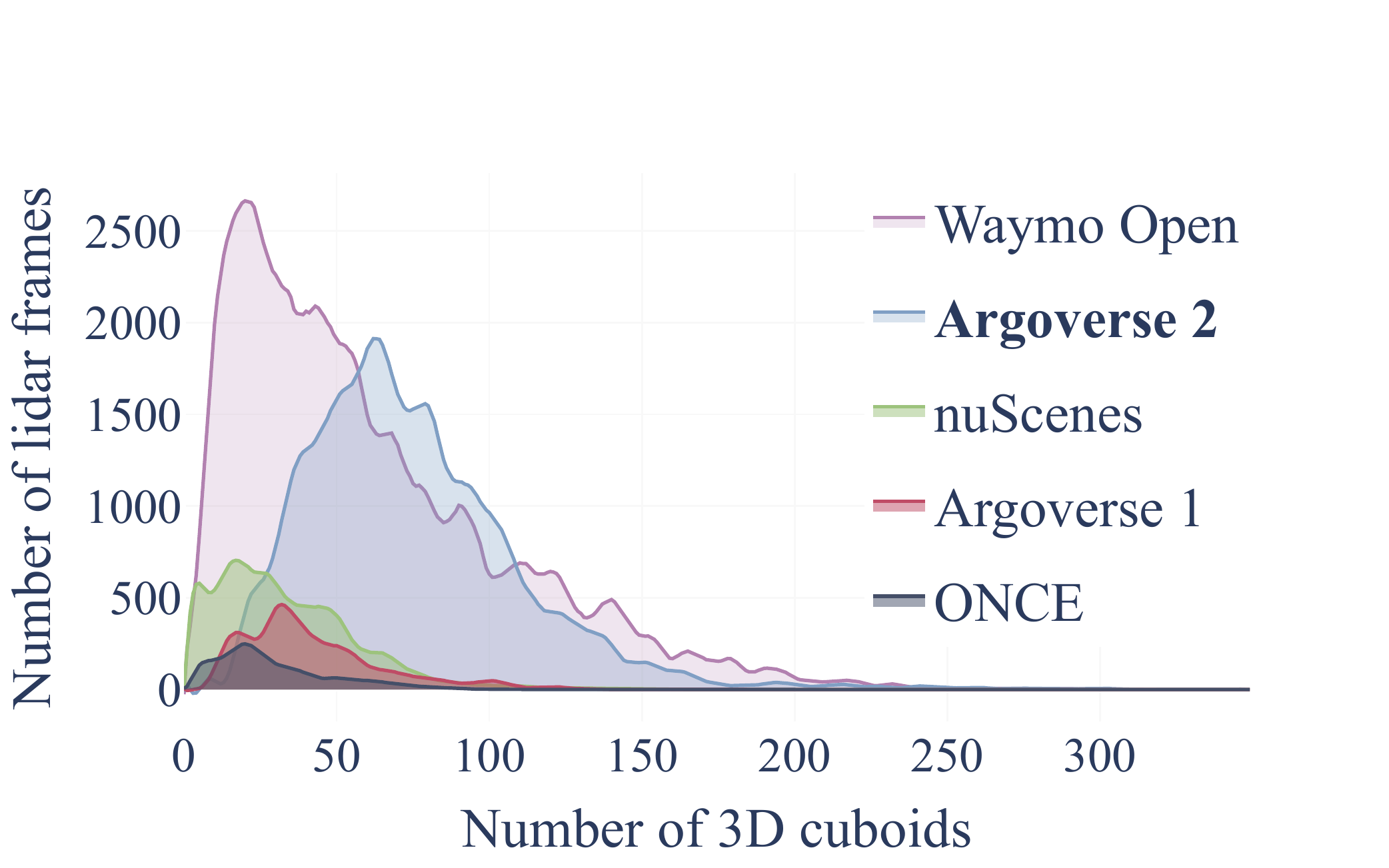}
    \caption{\textbf{Left:} Number of annotated 3D cuboids by range in the Argoverse 2 \emph{Sensor Dataset}. About 14\% of the Argoverse 2 cuboids are beyond 75 m -- Waymo Open, nuScenes, and ONCE have less than 1\%. \textbf{Right:} Number of 3D cuboids per lidar frame. Argoverse 2 has an average of 75 3D cuboids per lidar frame -- Waymo Open has an average of 61, nuScenes 33, and ONCE 30.}
    \label{fig:cuboids-stats-range}
\end{figure}

\begin{figure}
    \centering
    \includegraphics[width=0.45\linewidth]{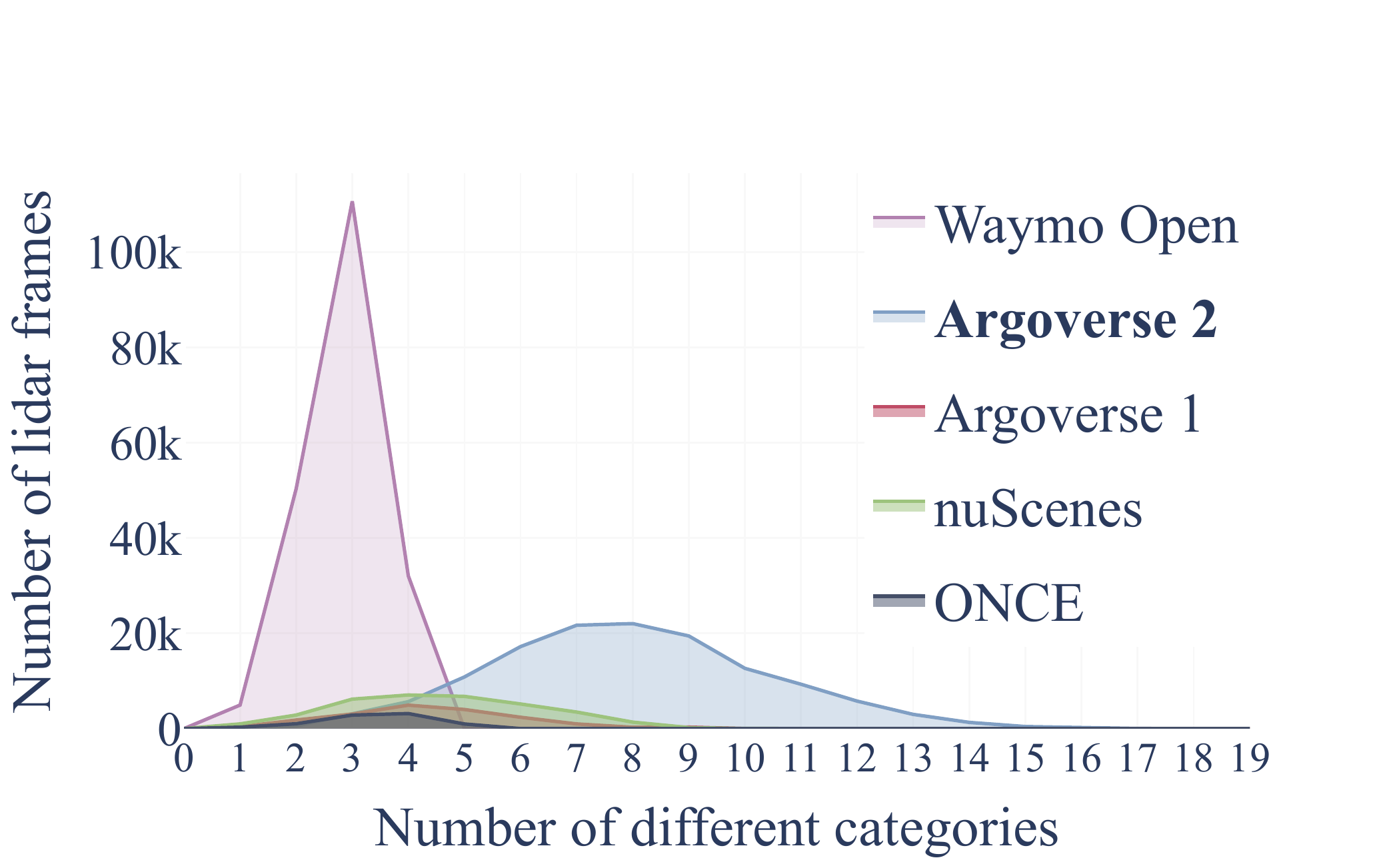}\qquad
    \includegraphics[width=0.45\linewidth]{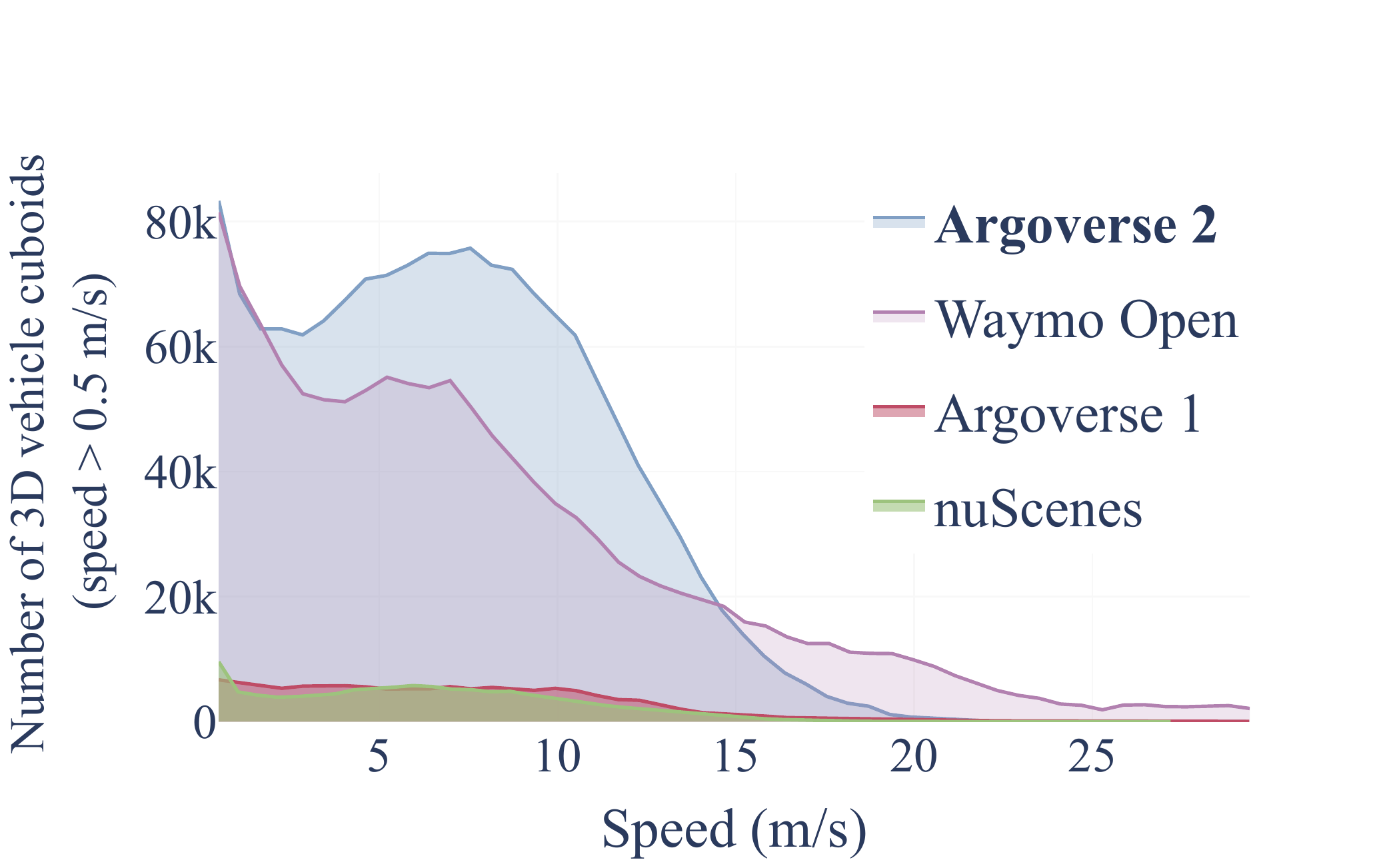}
    \caption{\textbf{Left:} Number of annotated categories per lidar frame in the Argoverse 2 \emph{Sensor Dataset}. Per scene, Argoverse 2 is about 2$\times$ more diverse than Argoverse 1 and 2.3$\times$ more diverse than Waymo Open. \textbf{Right:} Speed distribution for the vehicle category. We consider only moving vehicles with speeds greater than 0.5 m/s. Argoverse 2 has about 1.3$\times$ more moving vehicles than Waymo Open. About 28$\%$ of the vehicles in Argoverse 2 are moving with an average speed of 7.27 m/s. We did not compare against the ONCE dataset because it does not provide tracking information for the 3D cuboids.}
    \label{fig:cuboids-stats-speed}
\end{figure}

The \emph{Argoverse 2 Sensor Dataset} is the successor to the \emph{Argoverse 1 3D Tracking Dataset}. AV2 is larger, with 1,000 scenes, up from 113 in Argoverse 1, but each AV2 scene is also richer -- there are 23x as many non-vehicle, non-pedestrian cuboids in AV2. The constituent \SI{30}{\second} scenarios in the Argoverse 2 Sensor Dataset were manually selected by the authors to contain crowded scenes with under-represented objects, noteworthy weather, and interesting behaviors, e.g., cut ins and jaywalking. Each scenario is fifteen seconds in duration. Table~\ref{tab:sensor-datasets-comparison} compares the AV2 Sensor Dataset with a selection of self-driving datasets. Figures~\ref{fig:cuboids-stats},~\ref{fig:cuboids-stats-range}, and~\ref{fig:cuboids-stats-speed} plot how the scenarios of AV2 compare favorably to other datasets in terms of annotation range, object diversity, object density, and scene dynamism. 

The most similar sensor dataset to ours is the highly influential nuScenes~\cite{Caesar20cvpr_NuScenes} -- both datasets have 1,000 scenarios and HD maps, although Argoverse is unique in having ground height maps. nuScenes contains radar data while AV2 contains stereo imagery. nuScenes has a large taxonomy -- twenty-three object categories of which ten have suitable data for training and evaluation. Our dataset contains thirty object categories of which twenty-six are well sampled enough for training and evaluation. nuScenes spans two cities, while our proposed dataset spans six.

\input{neurips_data_2021/tables/dataset-comparison}

\paragraph{Sensor Suite.} \label{sensors} Lidar sweeps are collected at \SI{10}{\hertz}, along with \SI{20}{fps} imagery from 7 cameras positioned to provide a fully panoramic field of view. In addition, camera intrinsics, extrinsics and 6-DOF ego-vehicle pose in a global coordinate system are provided. Lidar returns are captured by two 32-beam lidars, spinning at \SI{10}{\hertz} in the same direction, but separated in orientation by \ang{180}. The cameras trigger in-sync with both lidars, leading to a \SI{20}{\hertz} frame-rate. The seven global shutter cameras are synchronized to the lidar to have their exposure centered on the lidar sweeping through their fields of view. 
In the Appendix, we provide a a schematic figure illustrating the car sensor suite and its coordinate frames. %

\textbf{Lidar synchronization accuracy.} In AV2, we improve the synchronization of cameras and lidars significantly over Argoverse 1. Our synchronization accuracy is within $[-1.39,1.39]$ \SI{}{ms}, which compares favorably to the Waymo Open Dataset, which is reported as $[-6,7]$ \SI{}{ms}~\cite{Sun20cvpr_WaymoOpenDataset}.

\textbf{Annotations.} The AV2 Sensor Dataset contains \SI{10}{Hz} 3D cuboid annotations for objects within our 30 class taxonomy (Figure~\ref{fig:cuboids-stats}). Cuboids have track identifiers that are consistent over time for the same object instance. Objects are annotated if they are within the ``region of interest'' (ROI) -- within five meters of the mapped ``driveable'' area.

\textbf{Privacy.} All faces and license plates, whether inside vehicles or outside of the driveable area, are blurred extensively to preserve privacy.
\label{privacy}

\textbf{Sensor Dataset splits}. We randomly partition the dataset with train, validation, and test splits of 700, 150, and 150 scenarios, respectively.

\subsection{Lidar Dataset}
The \emph{Argoverse 2 Lidar Dataset} is intended to support research into self-supervised learning in the lidar domain as well as point cloud forecasting~\cite{Weng20eccvw_SPF,Weng20corl_SPF2SequentialPointcloudForecasting}. Because lidar data is more compact than the full sensor suite, we can include double-length scenarios (\SI{30}{\second} instead of \SI{15}{\second}), and far more -- 20,000 instead of 1,000 -- equating to roughly 40x as many driving hours, for 5x the space budget. The AV2 Lidar Dataset is mined with the same criteria as the Forecasting Dataset (Section~\ref{mining}) to ensure that each scene is interesting. While the Lidar Dataset does not have 3D object annotations, each scenario carries an HD map with rich, 3D information about the scene.

Our dataset is the largest such collection to date with 20,000 thirty second sequences. The only similar dataset, concurrently released ONCE~\cite{Mao21arxiv_ONCEOneMillionScenes}, contains \SI{1}{M} lidar frames compared to \SI{6}{M} lidar frames in ours. Our dataset is sampled at \SI{10}{\hertz} instead of \SI{2}{\hertz}, as in ONCE, making our dataset more suitable for point cloud forecasting or self-supervision tasks where point cloud evolution over time is important. 

\textbf{Lidar Dataset splits}. We randomly partition the dataset with train, validation, and test splits of 16,000, 2,000, and 2,000 scenarios, respectively.

\subsection{Motion Forecasting Dataset}

\input{neurips_data_2021/tables/dataset-comparison-forecasting}

Motion forecasting addresses the problem of predicting future states (or occupancy maps) for dynamic actors within a local environment.  Some examples of relevant actors for autonomous driving include: vehicles (both parked and moving), pedestrians, cyclists, scooters, and pets.  Predicted futures generated by a forecasting system are consumed as the primary inputs in motion planning, which conditions trajectory selection on such forecasts. Generating these forecasts presents a complex, multi-modal problem involving many diverse, partially-observed, and socially interacting agents. However, by taking advantage of the ability to ``self-label'' data using observed ground truth futures, motion forecasting becomes an ideal domain for application of machine learning. 

Building upon the success of Argoverse 1, the Argoverse 2 Motion Forecasting dataset provides an updated set of prediction scenarios collected from a self-driving fleet. The design decisions enumerated below capture the collective lessons learned from both our internal research/development, as well as feedback from more than 2,700 submissions by nearly 260 unique teams\footnote{This count includes private submissions not posted to the public leaderboards.} across 3 competitions \cite{argoverse_forecasting_challenge}:

\begin{enumerate}[leftmargin=*]
    \item \textbf{Motion forecasting is a safety critical system in a long-tailed domain.}  Consequently, our dataset is biased towards diverse and interesting scenarios containing different types of focal agents (see section \ref{mining}).  Our goal is to encourage the development of methods that ensure safety during tail events, rather than to optimize the expected performance on ``easy miles''.
    \item \textbf{There is a ``Goldilocks zone'' of task difficulty.}  Performance on the Argoverse 1 test set has begun to plateau, as shown in Figure \ref{fig:mf_score_plateau} of the appendix.  Argoverse 2 is designed to increase prediction difficulty incrementally, spurring productive focused research for the next few years.  These changes are intended to incentivize methods that perform well on extended forecast horizons (3 s $\rightarrow$ 6 s), handle multiple types of dynamic objects (1 $\rightarrow$ 5), and ensure safety in scenarios from the long tail.  Future Argoverse releases could continue to increase the problem difficulty by reducing observation windows and increasing forecasting horizons.
    \item \textbf{Usability matters.} Argoverse 1 benefited from a large and active research community---in large part due to the simplicity of setup and usage.  Consequently, we took care to ensure that existing Argoverse models can be easily ported to run on Argoverse 2.  In particular, we have prioritized intuitive access to map elements,  encouraging methods which use the lane graph as a strong prior. To improve training and generalization, all poses have also been interpolated and resampled at exactly \SI{10}{\hertz} (Argoverse 1 was approximate). The new dataset includes fewer, but longer and more complex scenarios; this ensures that total dataset size remains large enough to train complex models but small enough to be readily accessible.
\end{enumerate}

\subsubsection{Data Representation}
\begin{figure}
    \centering
    \includegraphics[width=\columnwidth]{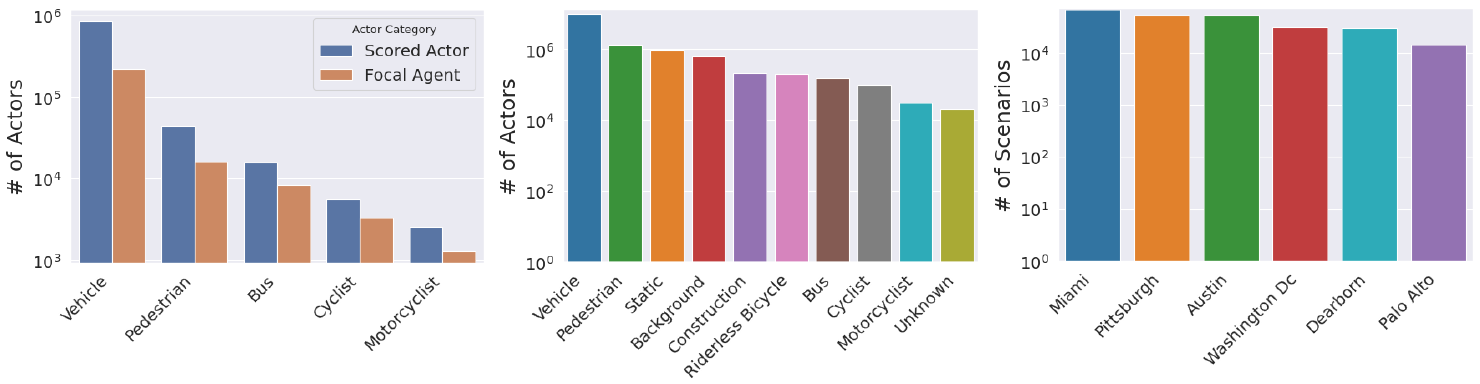}
    \caption{Object type and geographic histograms for the Motion Forecasting Dataset. \textbf{Left}: Histogram of object types over the ``focal'' and ``scored'' categories. \textbf{Center}: Histogram of object types over all tracks present in the dataset.   The fine grained distinctions between different static object types (e.g. \textit{Construction Cone} vs \textit{Riderless Bicycle}) are unique among forecasting datasets. \textbf{Right}: Histogram of metropolitan areas included in the dataset.}
    \label{fig:forecasting_track_and_metro_analysis}
\end{figure}
The dataset consists of 250,000 non-overlapping scenarios (80/10/10 train/val/test random splits) mined from six unique urban driving environments in the United States. It contains a total of 10 object types, with 5 from each of the dynamic and static categories (see Figure~\ref{fig:forecasting_track_and_metro_analysis}). Each scenario includes a local vector map and \SI{11}{\second} (\SI{10}{\hertz}) of trajectory data (2D position, velocity, and orientation) for all tracks observed by the ego-vehicle in the local environment. The first \SI{5}{\second} of each scenario is denoted as the \textit{observed} window, while the subsequent \SI{6}{\second} is denoted as the \textit{forecasted} horizon.

\begin{figure}
    \centering
    \includegraphics[width=\columnwidth]{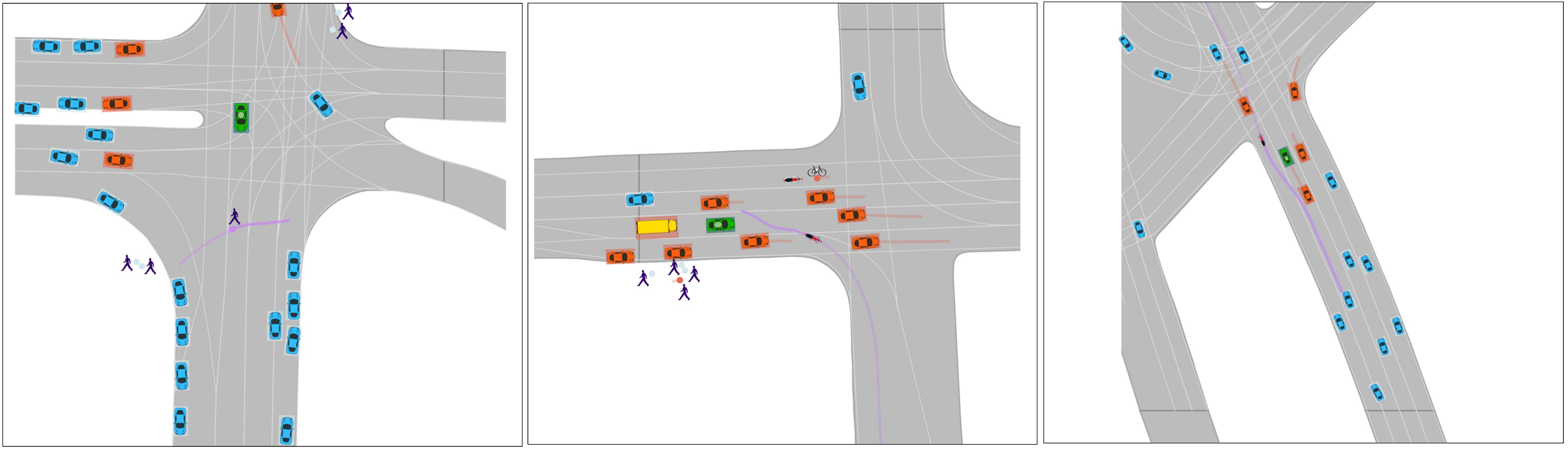}
    \caption{Visualization of a few interesting scenarios from the Motion Forecasting Dataset. The scenarios demonstrate a mix of the various object types (\textit{Vehicle}, \textit{Pedestrian}, \textit{Bus}, \textit{Cyclist}, or \textit{Motorcyclist}). The ego-vehicle is indicated in green, the focal agent is purple, and scored actors are orange.  Other un-scored tracks are shown in blue. Object positions are captured at the last timestep of the \textit{observed} history.  For visualization purposes the full \SI{5}{\second} history and \SI{6}{\second} future are rendered for the focal agent, while only \SI{1.5}{\second} of future are shown for the other scored actors. \textbf{Left} shows a pedestrian crossing in front of the ego-vehicle, while \textbf{center} and \textbf{right} depict a motorcyclist weaving through traffic. }
    \label{fig:multiverse_mockup}
\end{figure}

Within each scenario, we mark a single track as the ``focal agent''. Focal tracks are guaranteed to be fully observed throughout the duration of the scenario and have been specifically selected to maximize interesting interactions with map features and other nearby actors (see Section \ref{mining}). To evaluate multi-agent forecasting, we also mark a subset of tracks as ``scored actors'' (as shown in Figure \ref{fig:multiverse_mockup}), with guarantees for scenario relevance and minimum data quality. 

\subsubsection{Mining Interesting Scenarios}\label{mining}

The source data for Argoverse 2 was drawn from fleet logs tagged with annotations consistent with interesting or difficult-to-forecast events. Each log was trimmed to \SI{30}{\second} and run through an \textit{interestingness} scoring module in order to bias data selection towards examples from the long-tail of the natural distribution. We employ heuristics to score each track in the scene across five dimensions: object category, kinematics, map complexity, social context, and relation to the ego-vehicle (details in Appendix).

\begin{figure}[ht]
    \centering
    \includegraphics[width=0.45\columnwidth]{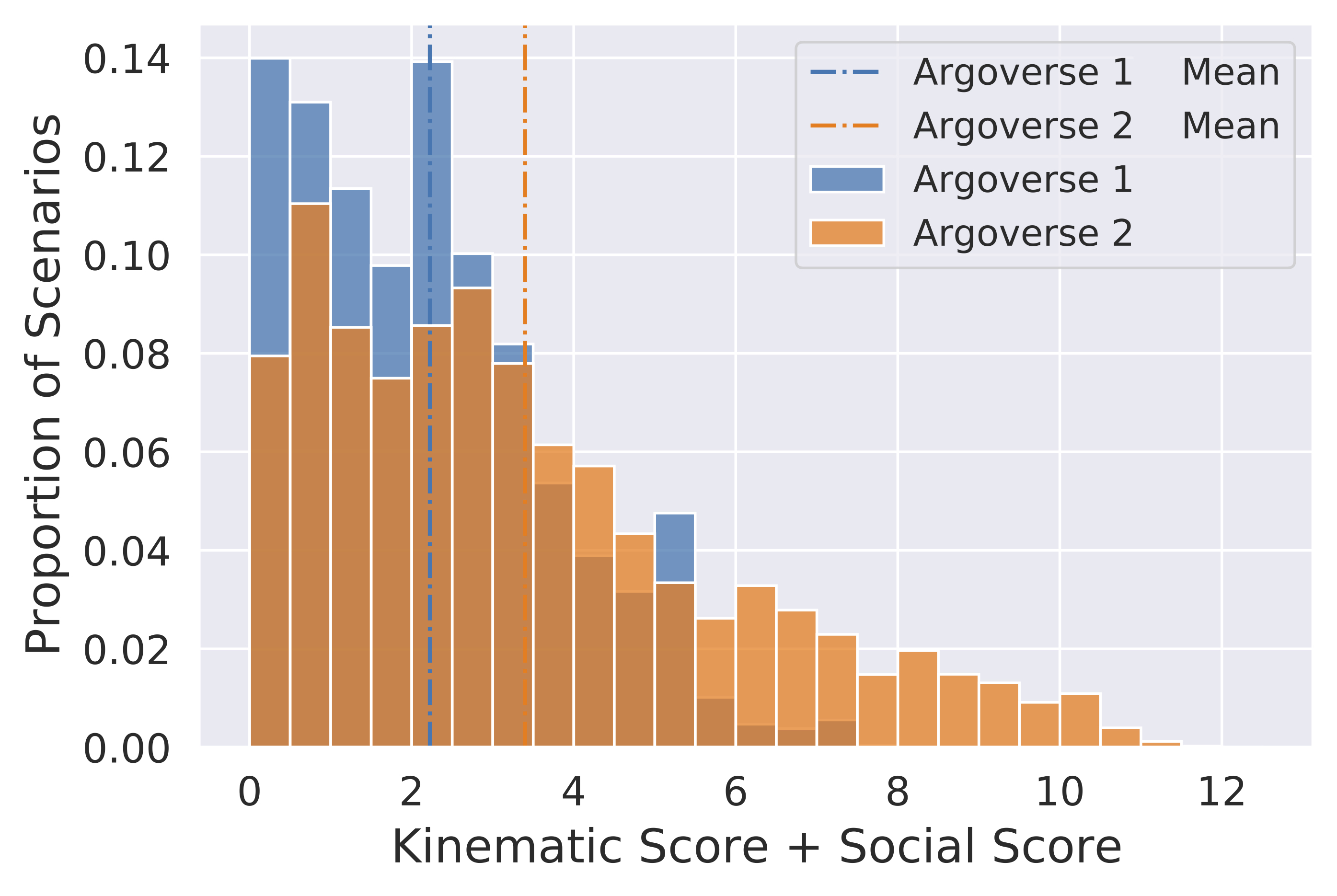}\qquad
    \includegraphics[width=0.45\columnwidth]{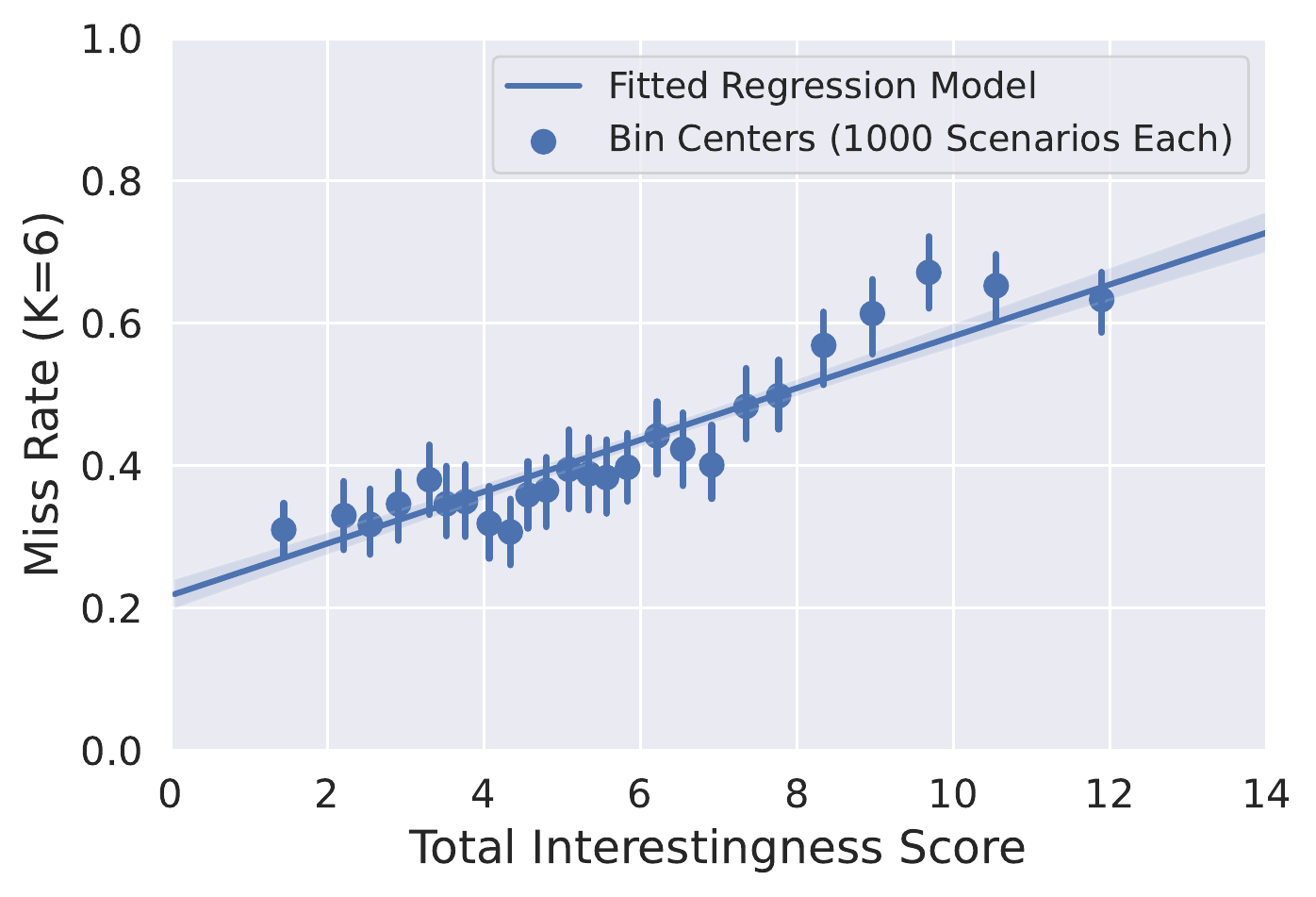}
    \caption{\textbf{Left:} Histogram comparing the distribution of interestingness scores assigned to focal agents in both Argoverse 1.1 and 2. \textbf{Right:} Plot showing the relationship between total \emph{interestingness score} and prediction difficulty on the Argoverse 2 test split. We evaluate WIMP \cite{khandelwal2020if} over each scenario and fit a regression model to the computed miss rate (K=6, 2m threshold).  }
    \label{fig:interestingness}
\end{figure}

The final scenarios are generated by extracting non-overlapping \SI{11} {\second} windows where at least one candidate track is fully observed for the entire duration.  The highest scoring candidate track is denoted as the ``focal agent'';  all other fully observed tracks within \SI{30}{\meter} of the ego-vehicle are denoted as ``scored actors''.  The resulting dataset is diverse, challenging, and still right-sized for widespread use (see the download size in Table \ref{forecasting-comparison}). In Figure \ref{fig:interestingness}, we show that the resulting dataset is significantly more interesting than Argoverse 1.1 and validate our intuition that actors scoring highly in our heuristic module are more challenging to accurately forecast.

\subsection{HD Maps}

Each scenario in the three datasets described above shares the same HD map representation. Each scenario carries its own local map region, similar to the Waymo Open Motion~\cite{Ettinger21arxiv_WaymoOpenMotion} dataset. This is a departure from the original Argoverse datasets in which all scenarios were localized onto two city-scale maps---one for Pittsburgh and one for Miami. In the Appendix, we provide examples. %
Advantages of per-scenario maps include more efficient queries and their ability to handle \emph{map changes}. A particular intersection might be observed multiple times in our datasets, and there could be changes to the lanes, crosswalks, or even ground height in that time. 

\textbf{Lane graph.} The core feature of the HD map is the lane graph, consisting of a graph $\mathcal{G}=(\mathcal{V},\mathcal{E})$, where $\mathcal{V}$ are individual lane segments. In the Appendix, we enumerate and define the attributes we provide for each lane segment. Unlike Argoverse 1, we provide the actual 3D lane boundaries, instead of only centerlines. However, our API provides code to quickly infer the centerlines at any desired sampling resolution. Polylines are quantized to \SI{1}{\centi\meter} resolution. Our representation is richer than nuScenes, which provides lane geometry only in 2D, not 3D. %

\textbf{Driveable area.} Instead of providing driveable area segmentation in a rasterized format, as we did in Argoverse 1, we release it in a vector format, i.e. as 3D polygons. This offers multiple advantages, chiefly in compression, allowing us to store separate maps for tens of thousands of scenarios, yet the raster format is still easily derivable. The polygon vertices are quantized to \SI{1}{\centi\meter} resolution.

\textbf{Ground surface height.} Only the sensor dataset includes a dense ground surface height map (although other datasets still have sparse 3D height information on polylines). Ground surface height is provided for areas within a \SI{5}{\meter} isocontour of the driveable area boundary, which we define as the \emph{region of interest} (ROI) \cite{Chang19cvpr_Argoverse}. We do so because the notion of ground surface height is ill-defined for the interior of buildings and interior of densely constructed city blocks, areas where ground vehicles cannot observe due to occlusion. The raster grid is quantized to a \SI{30}{\centi\meter} resolution, a higher resolution than the \SI{1}{\meter} resolution in Argoverse 1.

\textbf{Area of Local Maps.} Each scenario's local map includes all entities found within a \SI{100}{\meter} dilation in $l_2$-norm from the ego-vehicle trajectory.

%% file: neurips_data_2021/tables/dataset-comparison.tex
\begin{table}[t]
  \vspace{-2mm}
  \caption{Comparison of the Argoverse 2 \emph{Sensor} and \emph{Lidar} datasets with other sensor datasets.}
  \label{tab:sensor-datasets-comparison}
  \centering
  \begin{adjustbox}{width=\columnwidth}
    \begingroup
    \renewcommand{\arraystretch}{1.25} %
  \begin{tabular}{l r rcrcc rr}
    \toprule
    Name & \# Scenes & Cities & Lidar? & \# Cameras & Stereo & HD Maps? & \# Classes & \# Evaluated Classes \\
    \midrule
    Argoverse 1~\cite{Chang19cvpr_Argoverse} & 113 & 2 & \checkmark & 7 & \checkmark & \checkmark & 15 & 3 \\
    KITTI~\cite{Geiger12cvpr_KITTIAreWeReady} & 22 & 1 & \checkmark  & 2 & \checkmark & & 3 & 3\\
    nuScenes~\cite{Caesar20cvpr_NuScenes} & 1,000 & 2 & \checkmark & 6 & & \checkmark & 23 & 10 \\
    ONCE~\cite{Mao21arxiv_ONCEOneMillionScenes} & 581 & -- & \checkmark & 7 & & & 5 & 3 \\
    Waymo Open~\cite{Sun20cvpr_WaymoOpenDataset} & 1,150 & 3 & \checkmark & 5 & & & 4 & 4 \\
    \midrule
    Argoverse 2 Sensor & 1,000 & 6 & \checkmark & 9 & \checkmark & \checkmark & 30 & 26 \\
    Argoverse 2 Lidar &  20,000  & 6 & \checkmark & - & & \checkmark & - & - \\
    \bottomrule
  \end{tabular}
   \endgroup
  \end{adjustbox}
\end{table}

%% file: neurips_data_2021/tables/dataset-comparison-forecasting.tex
\begin{table}[t]
\vspace{-3mm}
  \caption{Comparison between the Argoverse 2 Motion Forecasting dataset and other recent motion forecasting datasets. Hyphens "-" indicate that attributes are either not applicable, or not available. We define ``mined for interestingness'' to be true if interesting scenarios/actors are mined \textit{after data collection}, instead of taking all/random samples. $\dagger$ Public leaderboard counts as retrieved on Aug. 27, 2021.}
  \label{forecasting-comparison}
  \centering
    \begin{adjustbox}{max width=\columnwidth}
    \begingroup
    \renewcommand{\arraystretch}{1.25} %
  \begin{tabular}{rccccccc}
    \toprule
    & \textsc{Argoverse}~\cite{Chang19cvpr_Argoverse} & \textsc{Inter}~\cite{zhan2019interaction} & \textsc{Lyft}~\cite{Houston20arxiv_LyftL5} & \textsc{Waymo}~\cite{Ettinger21arxiv_WaymoOpenMotion} & \textsc{NuScenes}~\cite{Caesar20cvpr_NuScenes} & \textsc{Yandex}~\cite{Malinin21arxiv_ShiftsYandex} & \textsc{Ours} \\
    \midrule
    \textsc{\# Scenarios} & 324k & - & 170k & 104k & 41k & 600k & 250k \\
    \textsc{\# Unique Tracks} & 11.7M  & 40k & 53.4M & 7.6M & - & 17.4M & 13.9M  \\
    \textsc{Average Track Length} & 2.48 s & 19.8 s & 1.8 s & 7.04 s & - & - & 5.16 s  \\
    \textsc{Total Time} & 320 h & 16.5 h & 1118 h & 574 h & 5.5 h & 1667 h & 763 h \\
    \textsc{Scenario Duration} & 5 s & - & 25 s & 9.1 s & 8 s & 10 s & 11 s \\
    \textsc{Test Forecast Horizon} & 3 s & 3 s & 5 s & 8 s & 6 s & 5 s & 6 s \\
    \textsc{Sampling Rate} & 10 Hz & 10 Hz & 10 Hz & 10 Hz & 2 Hz & 5 Hz & 10 Hz \\
    \textsc{\# Cities} & 2 & 6 & 1 & 6 & 2 & 6 & 6  \\
    \textsc{Unique Roadways} & 290 km & 2 km & 10 km & 1750 km & - & - & 2220 km \\
    \textsc{Avg. \# tracks per scenario} & 50 & - & 79 & - & 75 & 29 & 73\\
    \textsc{\# Evaluated object categories} & 1 & 1 & 3 & 3 & 1 & 2 & 5 \\
    \textsc{Multi-agent evaluation} & $\times$ & \checkmark & \checkmark & \checkmark & $\times$ & \checkmark &\checkmark \\
    \textsc{Mined for Interestingness} & \checkmark & $\times$ & - & \checkmark & $\times$ & $\times$ & \checkmark \\
    \textsc{Vector Map} & \checkmark & $\times$ & $\times$ & \checkmark & \checkmark & $\times$ &\checkmark \\
    \textsc{Download Size} & 4.8 GB  & -  & 22 GB  & 1.4 TB  & 48 GB & 120 GB & 58 GB \\
    \textsc{\# Public Leaderboard Entries}$^\dagger$ & 194  & -  & 935  & 23 & 18 & 3 & - \\
    \bottomrule
  \end{tabular}
   \endgroup
  \end{adjustbox}
  \vspace{-5mm}
\end{table}

%% file: neurips_data_2021/sections/evaluation.tex
\section{Experiments}

Argoverse 2 supports a variety of downstream tasks. In this section we highlight three different learning problems: 3D object detection, point cloud forecasting, and motion forecasting --- each supported by the sensor, lidar, and motion forecasting datasets, respectively.
First, we illustrate the \emph{challenging} and \emph{diverse} taxonomy within the Argoverse 2 sensor dataset by training a state-of-the-art 3D detection model on our twenty-six evaluation classes including ``long-tail'' classes such as stroller, wheel chairs, and dogs.
Second, we showcase the utility of the Argoverse 2 lidar dataset through \emph{large-scale}, self-supervised learning through the point cloud forecasting task.
Lastly, we demonstrate motion forecasting experiments which provide the first baseline for broad taxonomy motion prediction.

\subsection{3D Object Detection}

\input{neurips_data_2021/tables/sensor-dataset-3d-object-detection}

We provide baseline 3D detection results using a state-of-the-art, anchorless 3D object detection model -- CenterPoint~\cite{Yin21cvpr_CenterPoint}.
Our CenterPoint implementation takes a point cloud as input and crops it to a \SI{200}{\meter} $\times$ \SI{200}{\meter} grid with a voxel resolution of [\SI{0.1}{\meter},  \SI{0.1}{\meter}] in the \(xy\) (bird's-eye-view) plane and \SI{0.2}{\meter} in the \(z\)-axis.
To accommodate our larger taxonomy, we include six detection heads to encourage feature specialization. Figure \ref{fig:3d-detection-baseline} characterizes the performance of our 3D detection baseline using the nuScenes~ \cite{Caesar20cvpr_NuScenes} average precision metric. Our large taxonomy allows us to evaluate classes such as ``Wheeled Device'' (e-Scooter), ``Stroller'', ``Dog'', and ``Wheelchair'' and we find that performance on these categories with strong baselines is poor despite significant amounts of training data.

In Table \ref{tab:3d-obj-det-leaderboard-results}, we provide a snapshot of submissions to the Argoverse 2 3D Object Detection Leaderboard.

\subsection{Point Cloud Forecasting}

We perform point cloud forecasting according to the experimental protocol of SPF2~\cite{Weng20corl_SPF2SequentialPointcloudForecasting} using the Argoverse 2 Lidar Dataset. Given a sequence of past scene point clouds, a model is required to predict a sequence of future scene point clouds.
We take the scene point clouds in the past \SI{1}{\second} (\SI{10}{\hertz}) in the range image format as input, and then predict the next \SI{1}{\second} of range images. SPFNet predicts two output maps at each time step -- the first output map is the predicted range values, while the second output is a validity mask. %
Previous point cloud forecasting models were evaluated on smaller datasets such as KITTI or nuScenes. To explore how the amount of training data affects the performance, we use increasing amounts of data for training the same model architecture, up to the full training set of 16,000 sequences.

\paragraph{Evaluation.} We use three metrics to evaluate the performance of our forecasting model: \emph{mean IoU}, \emph{$l_1$-norm}, and \emph{Chamfer distance}. The \emph{mean IoU} evaluates the predicted range mask.
The \emph{$l_1$-norm} measures the average $l_1$ distance between the pixel sets of predicted range image and the ground-truth image, which are both masked out by the ground-truth range mask. The \emph{Chamfer distance} is obtained by adding up the Chamfer distances in both directions (forward and backward) between the ground-truth point cloud and the predicted scene point cloud which is obtained by back-projecting the predicted range image. %

\input{neurips_data_2021/tables/evaluation-point-forecast}

\paragraph{Results of SPF2 and Discussion.} Table~\ref{eval-point-forecast-baseline} contains the results of our point cloud forecasting experiments. With increasing training data, the performance of the model grows steadily in all three metrics. %
These results and the works from the self-supervised learning literature~\cite{brown2020language, chen2020simple} indicate that a large amount of training data can make a substantial difference. Another observation is that the Chamfer distances for predictions on our dataset are significantly higher than predictions on KITTI~\cite{Weng20corl_SPF2SequentialPointcloudForecasting}. We conjecture that this could be due to two reasons: (1) the Argoverse 2 Lidar Dataset has a much larger sensing range (above \SI{200}{\meter} versus \SI{120}{\meter} of the KITTI lidar sensor), which tends to significantly increase the value of Chamfer distance. (2) the Argoverse 2 Lidar Dataset has a higher proportion of dynamic scenes
compared with KITTI Dataset.%

\subsection{Motion Forecasting}
We present several forecasting baselines \cite{Chang19cvpr_Argoverse} which try to make use of different aspects of the data. Those which are trained using the focal agent only and do not capture any social interaction include: constant velocity, nearest neighbor, and LSTM encoder-decoder models (both with and without a map-prior). We also evaluate WIMP \cite{khandelwal2020if} as an example of a graph-based attention method that captures social interaction. All hyper-parameters are obtained from the reference implementations.

\paragraph{Evaluation.} Baseline approaches are evaluated according to standard metrics. Following \cite{Chang19cvpr_Argoverse}, we use \textit{minADE} and \textit{minFDE} as the metrics; they evaluate the average and endpoint L2 distance respectively, between the best forecasted trajectory and the ground truth. We also use \textit{Miss Rate (MR)} which represents the proportion of test samples where none of the forecasted trajectories were within 2.0 meters of ground truth according to endpoint error. The resulting performance illustrates both the community's progress on the problem and the significant increase in dataset difficulty when compared with Argoverse 1.1.

\input{neurips_data_2021/tables/evaluation-motion_forecasting_baselines}

\textbf{Baseline Results.} Table \ref{eval-mf-baseline} summarizes the results of baselines. 
For K=1, Argoverse 1 \cite{Chang19cvpr_Argoverse} showed that a constant velocity model (\textit{minFDE}=7.89) performed better than NN+map(prior) (\textit{minFDE}=8.12), which is not the case here. This further proves that Argoverse 2 is kinematically more diverse and cannot be solved by making constant velocity assumptions. Surprisingly, NN and LSTM variants that make use of a map prior perform worse than those which do not, illustrating the scope of improvement in how these baselines leverage the map. For K=6, WIMP significantly outperforms every other baseline. This emphasizes that it is imperative to train expressive models that can leverage map prior and social context along with making diverse predictions. The trends are similar to our past 3 Argoverse Motion Forecasting competitions  \cite{argoverse_forecasting_challenge}: Graph-based attention methods (e.g.  \cite{khandelwal2020if, Liang20eccv_LaneGCN, Mercat20icra_MultiheadAttentionForecasting}) continued to dominate the competition, and were nearly twice as accurate as the next best baseline (Nearest Neighbor) at K=6. That said, some of the rasterization-based (e.g. \cite{gilles2021home}) methods also showed promising results. Finally, we also evaluated baseline methods in the context of transfer learning and varied object types, the results of which are summarized in the Appendix.

In Table \ref{tab:mf-leaderboard-results}, we provide a snapshot of submissions to the Argoverse 2 Motion Forecasting Leaderboard.

%% file: neurips_data_2021/tables/sensor-dataset-3d-object-detection.tex
\begin{wrapfigure}{r}{0.5\textwidth}
\vspace{-2mm}
    \includegraphics[width=0.5\textwidth]{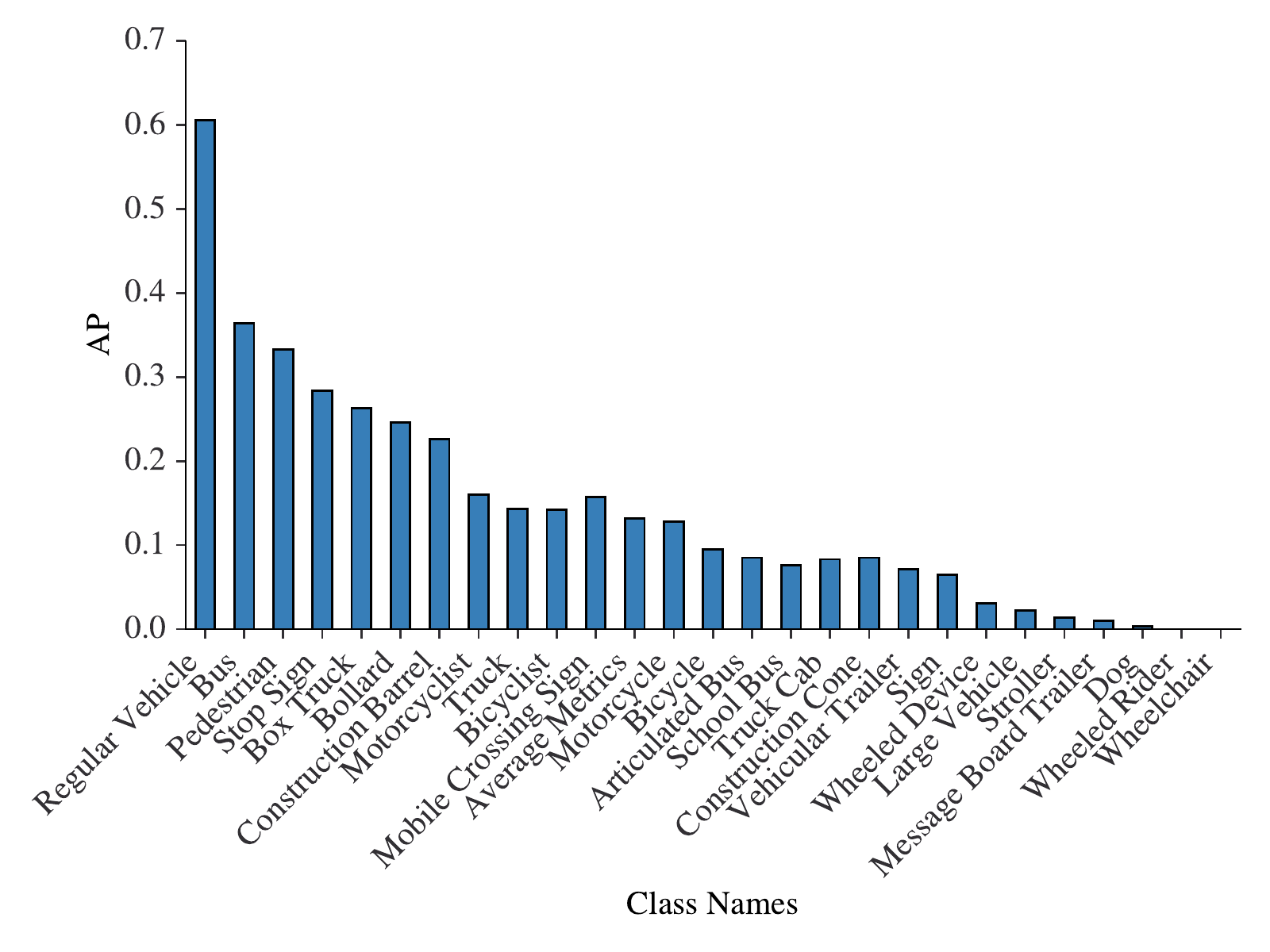}
    \caption{Average precision of our 3D object detection baseline on the \emph{validation} split of the \emph{Sensor Dataset (Beta)}. Our experiments showcase both our \emph{diverse} taxonomy and \emph{difficult} ``long-tail'' classes.}
    \label{fig:3d-detection-baseline}
    \vspace{-5mm}
\end{wrapfigure}

\begin{table*}
\caption{3d object detection results on the Argoverse 2 Sensor Dataset, taken from the \href{https://eval.ai/web/challenges/challenge-page/1710}{leaderboard} on Dec 21, 2022. \emph{Detectors} is the winner of the CVPR 2022 \emph{Workshop on Autonomous Driving} Argoverse 2 3D Object Detection challenge. }
\vspace{-2mm}
  \centering
    \begin{adjustbox}{max width=0.7\columnwidth}
    \begingroup
    \renewcommand{\arraystretch}{1.25} %
    \begin{tabular}{c| ccccc}
                               \textsc{Method}          & \textsc{mCDS} ($\uparrow$) & \textsc{mAP} ($\uparrow$) & \textsc{mATE} ($\downarrow$) & \textsc{mASE} ($\downarrow$) & \textsc{mAOE} ($\downarrow$) \\
   \toprule
   \textsc{CenterPoint (Ours)}                    & 0.14    & 0.18   & 0.49    & 0.34    & 0.72 \\
   \textsc{Detectors} \cite{Fang22tr_Detectors}   & 0.34    & 0.41   & \textbf{0.40}    & \textbf{0.30}    & 0.54 \\
   \textsc{BEVFusion} \cite{Liu22arxiv_BEVFusion} & \textbf{0.37}    & \textbf{0.46}   & \textbf{0.40}    & \textbf{0.30}    & \textbf{0.50} \\
   \bottomrule
  \end{tabular}
  \endgroup
  \end{adjustbox}
    \label{tab:3d-obj-det-leaderboard-results}
\end{table*}

%% file: neurips_data_2021/tables/evaluation-point-forecast.tex
\begin{table*}[b]
\caption{Results of point cloud forecasting on the \emph{test} split of the \emph{Lidar Dataset}.}
\vspace{-1mm}
  \label{eval-point-forecast-baseline}
  \centering
    \begin{adjustbox}{width=0.55\columnwidth}
    \begingroup
    \renewcommand{\arraystretch}{1.25} %
  \begin{tabular}{lrrrrrrr}
    \toprule
    & \multicolumn{7}{c}{\textsc{\# Train Logs}} \\
     & 125 & 250 & 500 & 1k & 2k & 4k & 16k  \\
    \midrule
    \textsc{mean IoU} (\%) ($\uparrow$) & 55.5 & 63.4 & 61.7 & 65.1 & 68.0 & 68.4 & \textbf{70.9} \\
    $l_1$\textsc{-norm} ($\downarrow$) & 13.5 & 12.5 & 11.8 & 9.9 & 8.9 & 8.7 & \textbf{7.4} \\
    \textsc{Chamfer dist.} ($\downarrow$) & 31.1 & 25.9 & 22.4 & 22.9 & 20.5 & 18.2 & \textbf{14.0} \\
    \bottomrule
  \end{tabular}
  \endgroup
  \end{adjustbox}
\end{table*}

%% file: neurips_data_2021/tables/evaluation-motion_forecasting_baselines.tex
\begin{table}[h]
  \caption{Performance of motion forecasting baseline methods on vehicle-like (\emph{vehicle}, \emph{bus}, \emph{motorcyclist}) object types from the \emph{Argoverse 2 Motion Forecasting (Beta)} Dataset. Usage of map prior indicates access to map information whereas usage of social context entails encoding other actors' states in the feature representation. Mining intersection (multimodal) scenarios leads to poor performance at K=1 for all methods.  Constant Velocity models have particularly poor performance due to the dataset bias towards kinematically interesting trajectories.  Note that modern deep methods such as WIMP still have a miss rate of 0.42 at K=6, indicating the increased difficulty of the Argoverse 2 dataset. Numbers within 1\% of the best are in bold.}
  \label{eval-mf-baseline}
  \centering
    \begin{adjustbox}{max width=\columnwidth}
    \begingroup
    \renewcommand{\arraystretch}{1.25} %
  \begin{tabular}{rcc|ccc|ccc}
    \toprule
    & & & \multicolumn{3}{c}{\textsc{K=1} } & \multicolumn{3}{c}{\textsc{K=6} } \\
    \textsc{Model} & \textsc{Map Prior} & \textsc{Social Context} & \textsc{minADE} $\downarrow$ & \textsc{minFDE} $\downarrow$ & \textsc{MR} $\downarrow$ & \textsc{minADE} $\downarrow$ & \textsc{minFDE} $\downarrow$ & \textsc{MR} $\downarrow$ \\
    \midrule
    \textsc{Const. Vel.}~\cite{Chang19cvpr_Argoverse} & & & 7.75 & 17.44  & 0.89 & - & - & - \\
    \textsc{NN}~\cite{Chang19cvpr_Argoverse} & & & 4.46 & 11.71 & \textbf{0.81} & 2.18 & 4.94 & 0.60 \\
    \textsc{NN}~\cite{Chang19cvpr_Argoverse}  & \checkmark & & 6.45 & 15.51 & 0.84 & 4.30 & 10.08 & 0.78 \\
    \textsc{LSTM}~\cite{Chang19cvpr_Argoverse} & & & \textbf{3.05} & 8.28 & 0.85 & - & - & - \\
    \textsc{LSTM}~\cite{Chang19cvpr_Argoverse} & \checkmark & & 5.07 & 12.71 & 0.90 & 3.73 & 9.09 & 0.85 \\
    \textsc{WIMP}~\cite{khandelwal2020if} & \checkmark & \checkmark & \textbf{3.09} & \textbf{7.71} & 0.84 & \textbf{1.47} & \textbf{2.90} & \textbf{0.42} \\ %
    \bottomrule
  \end{tabular}
  \endgroup
  \end{adjustbox}
  %
\end{table}

\begin{table}[h]
\caption{Motion forecasting results on the Argoverse 2 Motion Forecasting Dataset, taken from the online \href{https://eval.ai/web/challenges/challenge-page/1719}{leaderboard} on Dec 21, 2022. \emph{BANet} is the winner of the CVPR 2022 \emph{Workshop on Autonomous Driving} Argoverse 2 Motion Forecasting challenge (\#1), and QML and GANet received honorable mention (HM) prizes. Entries are sorted below according to \textit{Brier-minFDE}.}
\vspace{-1mm}
  \centering
    \begin{adjustbox}{max width=\columnwidth}
    \begingroup
    \renewcommand{\arraystretch}{1.25} %
    \begin{tabular}{c|ccc | cccc}
    \toprule
    &  \multicolumn{3}{c}{\textsc{K=1} } & \multicolumn{4}{c}{\textsc{K=6} } \\

    \textsc{Method} & \textsc{minADE} $\downarrow$ & \textsc{minFDE} $\downarrow$ & \textsc{MR} $\downarrow$ & \textsc{minADE} $\downarrow$ & \textsc{minFDE} $\downarrow$ & \textsc{MR} $\downarrow$ &	\textsc{brier-minFDE} $\downarrow$  \\
    \midrule
    \textsc{THOMAS (GOHOME Scalar)} \cite{Gilles22iclr_THOMAS} & 1.95 & 4.71 & 0.64 & 0.88 & 1.51 & 0.20  & 2.16 \\
    \textsc{GoRela (w/o ensemble)} \cite{Cui22arxiv_GoRela} & 1.82 & 4.62 & 0.61 & 0.76 & 1.48 & 0.22 & 2.01 \\
    \textsc{GANet (ensemble)} (HM)  \cite{Wang22arxiv_GANet}              & 1.81 & 4.57 & 0.61 & 0.73 & 1.36 & \textbf{0.17} & 1.98 \\
    \textsc{GANet (w/o ensemble)} \cite{Wang22arxiv_GANet} & \textbf{1.77} & \textbf{4.48} & \textbf{0.59} & 0.72 & \textbf{1.34} & \textbf{0.17} & 1.96 \\
    \textsc{QML} (HM) \cite{Su22tr_QML} & 1.84 & 4.98 & 0.62 & \textbf{0.69} & 1.39 & 0.19 & 1.95 \\
    \textsc{BANet (OPPred)} (\#1) \cite{Zhang22tr_BANet} & 1.79 & 4.61 & 0.60 & 0.71 & 1.36 & 0.19 & \textbf{1.92} \\

    \bottomrule
    \end{tabular}
  \endgroup
  \end{adjustbox}
    \label{tab:mf-leaderboard-results}
\end{table}

%% file: neurips_data_2021/sections/conclusion.tex
\section{Conclusion}

\textbf{Discussion.} In this work, we have introduced three new datasets that constitute Argoverse 2. We provide baseline explorations for three tasks -- 3d object detection, point cloud forecasting and motion forecasting. Our datasets provide new opportunities for many other tasks. We believe our datasets compare favorably to existing datasets, with HD maps, rich taxonomies, geographic diversity, and interesting scenes. 

\textbf{Limitations.} As in any human annotated dataset, there is label noise, although we seek to minimize it before release. 3D bounding boxes of objects are not included in the motion forecasting dataset, but one can make reasonable assumptions about the object extent given the object type. The motion forecasting dataset also has imperfect tracking, consistent with state-of-the-art 3D trackers.

%% file: neurips_data_2021/sections/supplementary.tex
\section{Appendix}

\subsection{Additional Information About Sensor Suite}
\begin{figure}[h]
    \centering
    \includegraphics[width=0.49\linewidth]{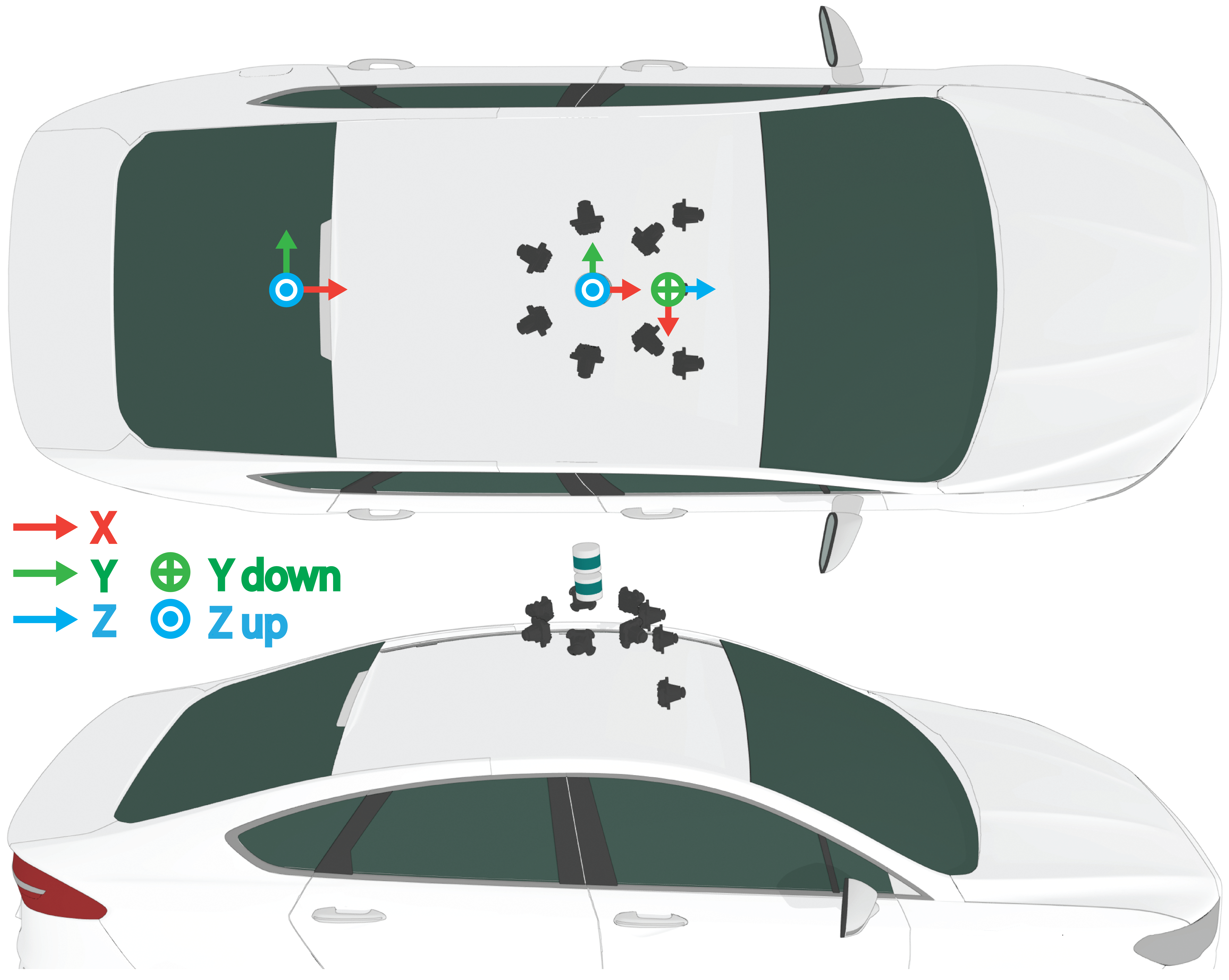}
    \caption{Car sensor schematic showing the three coordinate systems: (1) the vehicle frame in the rear axle; (2) the camera frame; and the lidar frame.}
    \label{fig:sensors}
\end{figure}

\begin{figure}[htbp]
    \centering
    \includegraphics[width=0.6\linewidth]{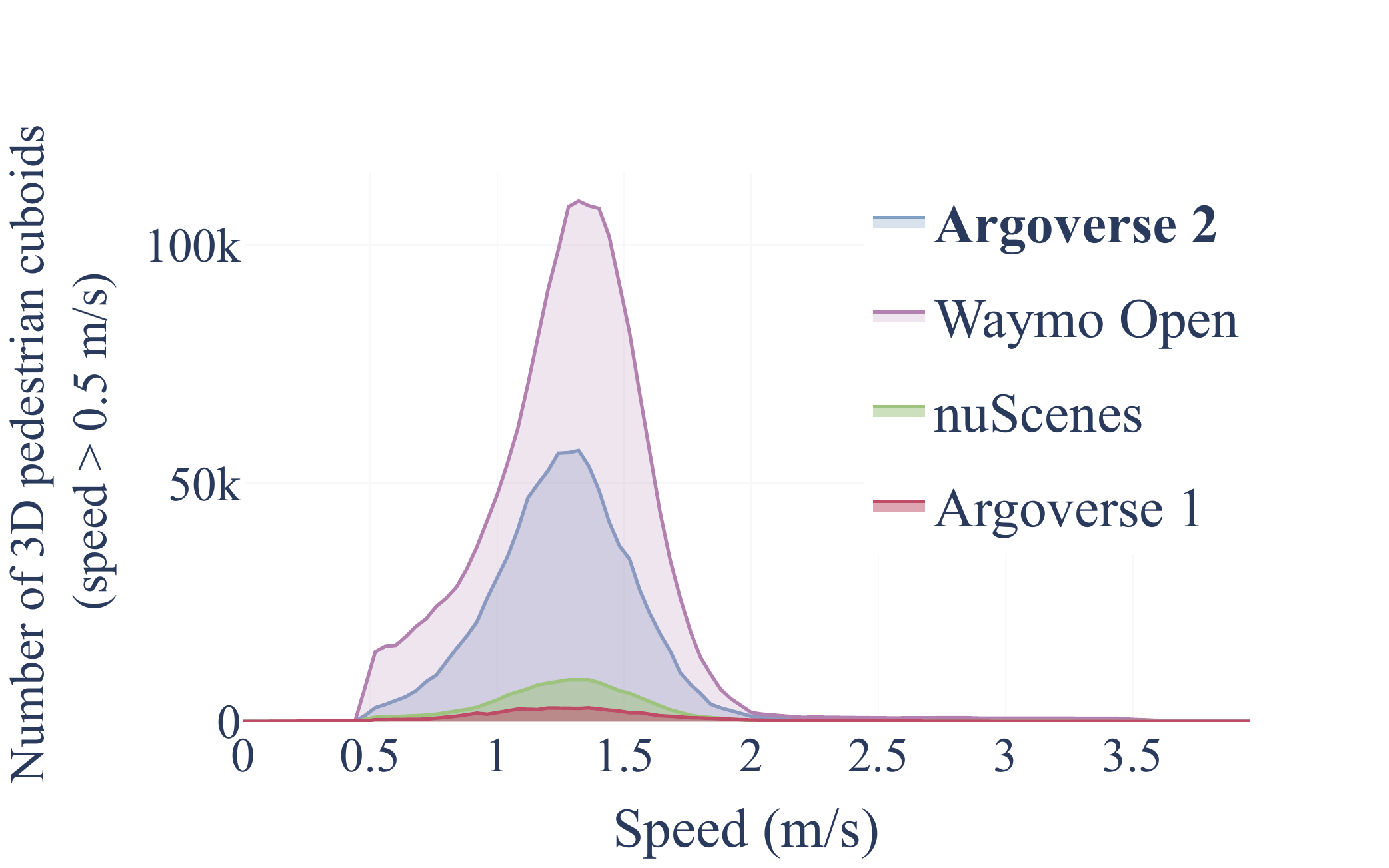}~~
    \includegraphics[width=0.40\linewidth]{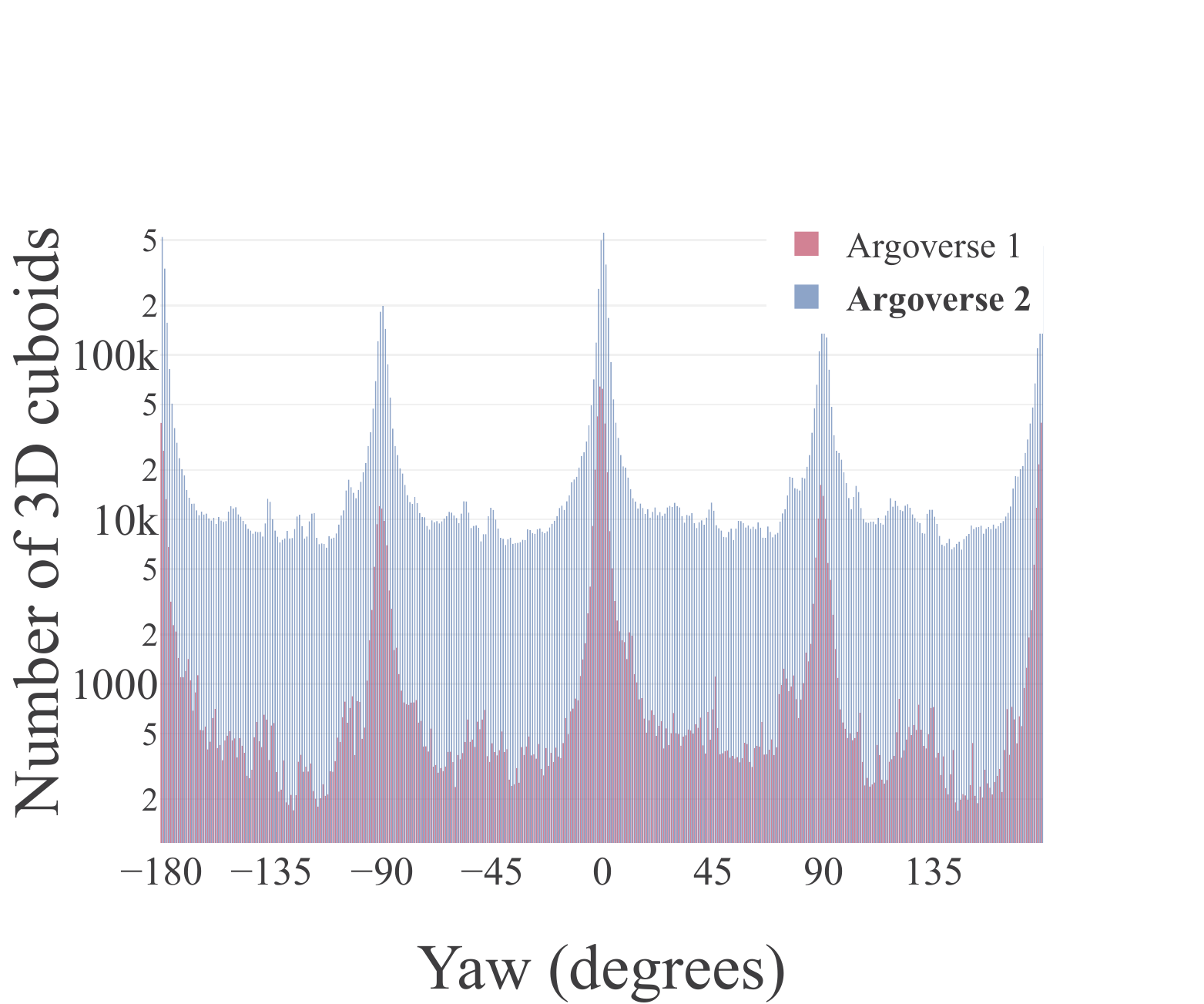}~~
    \caption{\textbf{Left}: Number of moving 3D cuboids for pedestrians by speed distribution. We define moving objects when the speed is greater than 0.5 m/s. \textbf{Right}: Number of annotated 3D cuboids by yaw distribution.}
    \label{fig:yaw-and-pedestrian-speed}
\end{figure}

In Figure~\ref{fig:sensors}, we provide a diagram of the sensor suite used to capture the Argoverse 2 datasets. Figure~\ref{fig:yaw-and-pedestrian-speed} shows the speed distribution for annotated pedestrian 3D cuboids and the yaw distribution.

\subsection{Additional Information About Motion Forecasting Dataset}

\subsubsection{Interestingness Scores}
Kinematic scoring selects for trajectories performing sharp turns or significant (de)accelerations.  The map complexity program biases the data set towards trajectories complex traversals of the underlying lane graph.  In particular, complex map regions, paths through intersections, and lane-changes score highly. Social scoring rewards tracks through dense regions of other actors.  Social scoring also selects for non-vehicle object classes to ensure adequate samples from rare classes, such as motorcycles, for training and evaluation.  Finally, the autonomous vehicle scoring program encourages the selection of tracks that intersect the ego-vehicle's desired route.

\begin{figure}[htbp]
    \centering
    \includegraphics[width=0.8\columnwidth]{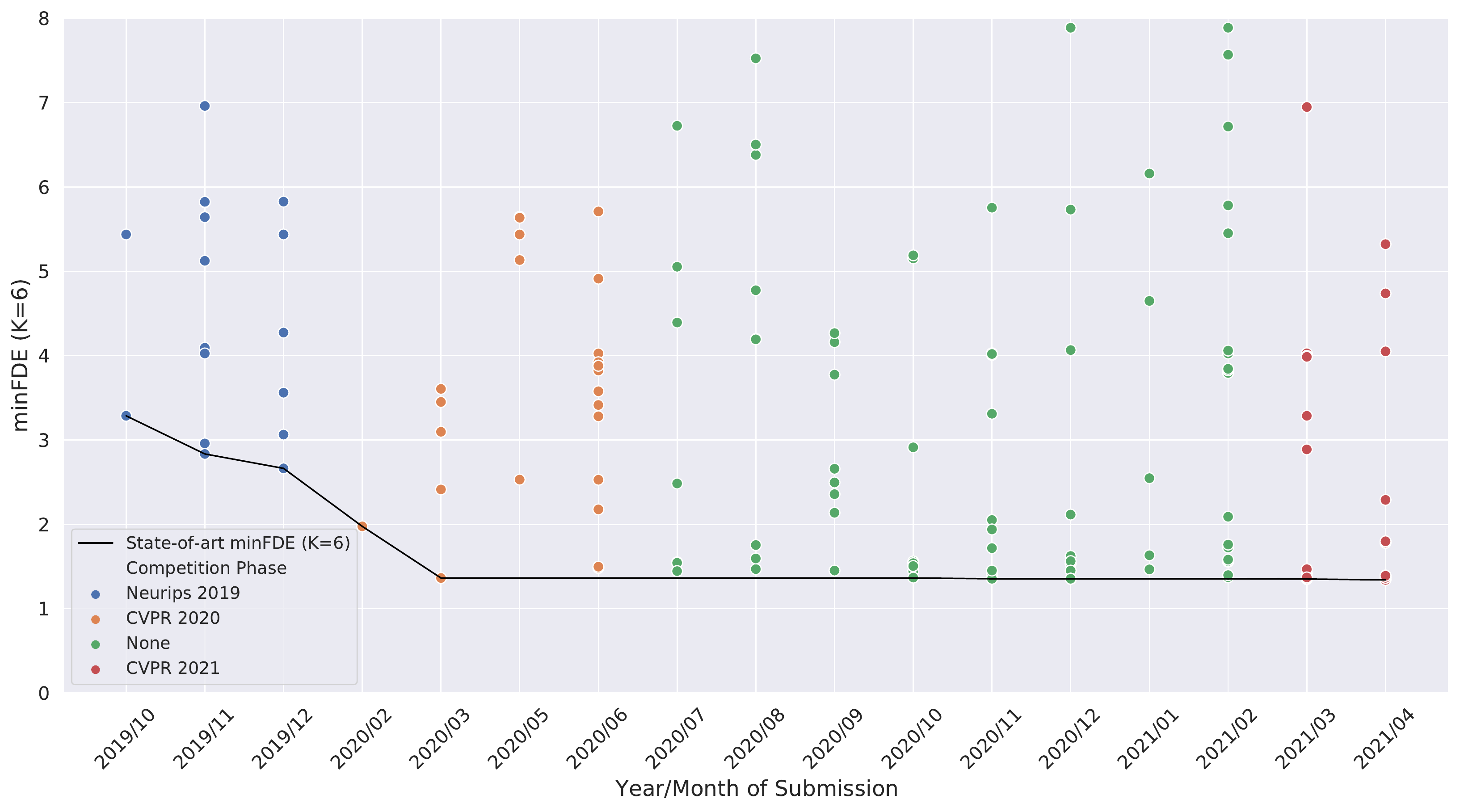}
    \caption{MinFDE metric values for submissions on Argoverse 1.1 over time.  Individual points indicate submissions to the public leader board.  Colors indicate specific competition phases.  The solid black line indicates SOTA performance.  The research community made massive gains which have plateaued since early 2020.  However, we note that the number and diversity of methods performing at or near the SOTA continues to grow. Additionally, later competitions sorted the leaderboard by ``Miss Rate'' and probability weighted FDE, and those metrics showed progress. Still, minFDE did not improve significantly.}
    \label{fig:mf_score_plateau}
\end{figure}

\begin{figure}[htbp]
    \centering
    \includegraphics[width=\columnwidth]{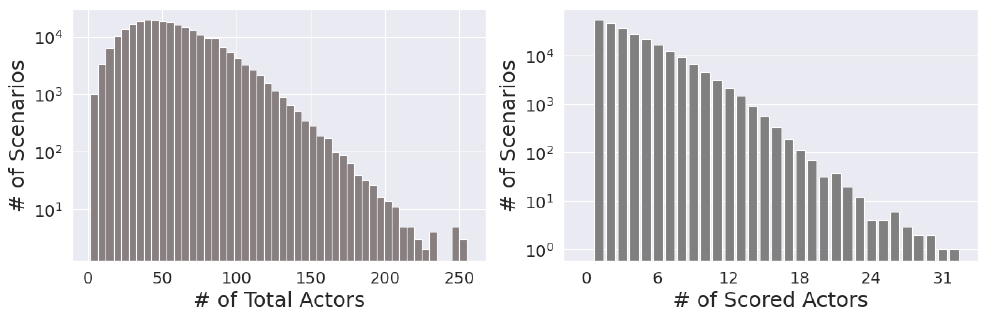}
    \caption{Histogram of the number of actors (both scored and all types) present in the Motion Forecasting Dataset scenarios. The Lidar Dataset is mined by the same criteria and thus follows the same distribution.}
    \label{fig:forecasting_num_tracks_analysis}
\end{figure}

\subsection{Additional Information About HD Maps}

\input{neurips_data_2021/tables/hd-map-attributes}

\paragraph{Examples of HD maps from the Sensor Dataset}
In Figure \ref{fig:sensordatasetmapexamples}, we display examples of local HD maps associated with individual logs/scenarios.

\input{neurips_data_2021/tables/sensor-dataset-map-examples-with-crosswalk}

\subsection{Additional 3D Detection Results}

\begin{figure}[htbp]
    \centering
    \includegraphics[scale=0.4]{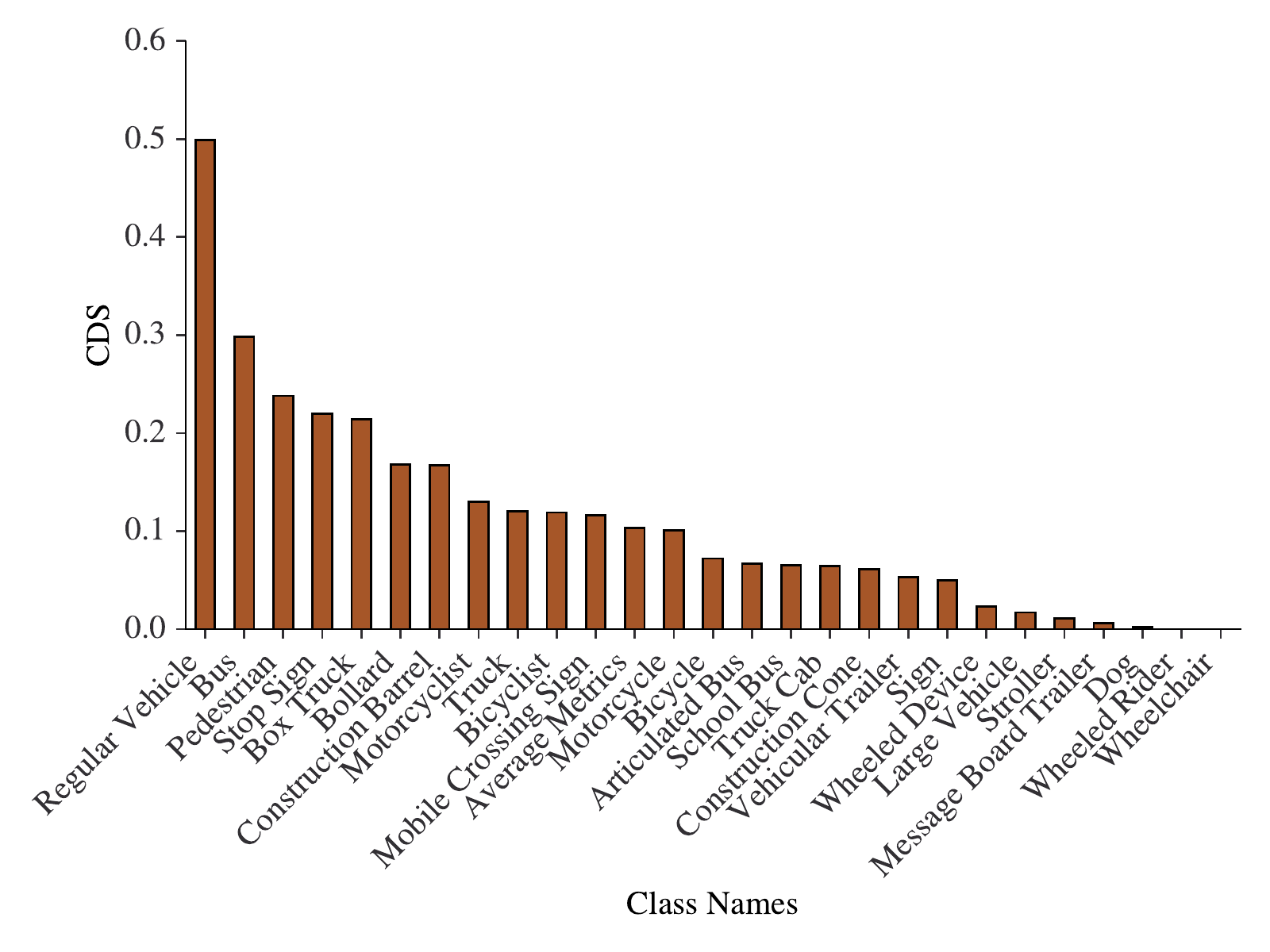}
    \includegraphics[scale=0.4]{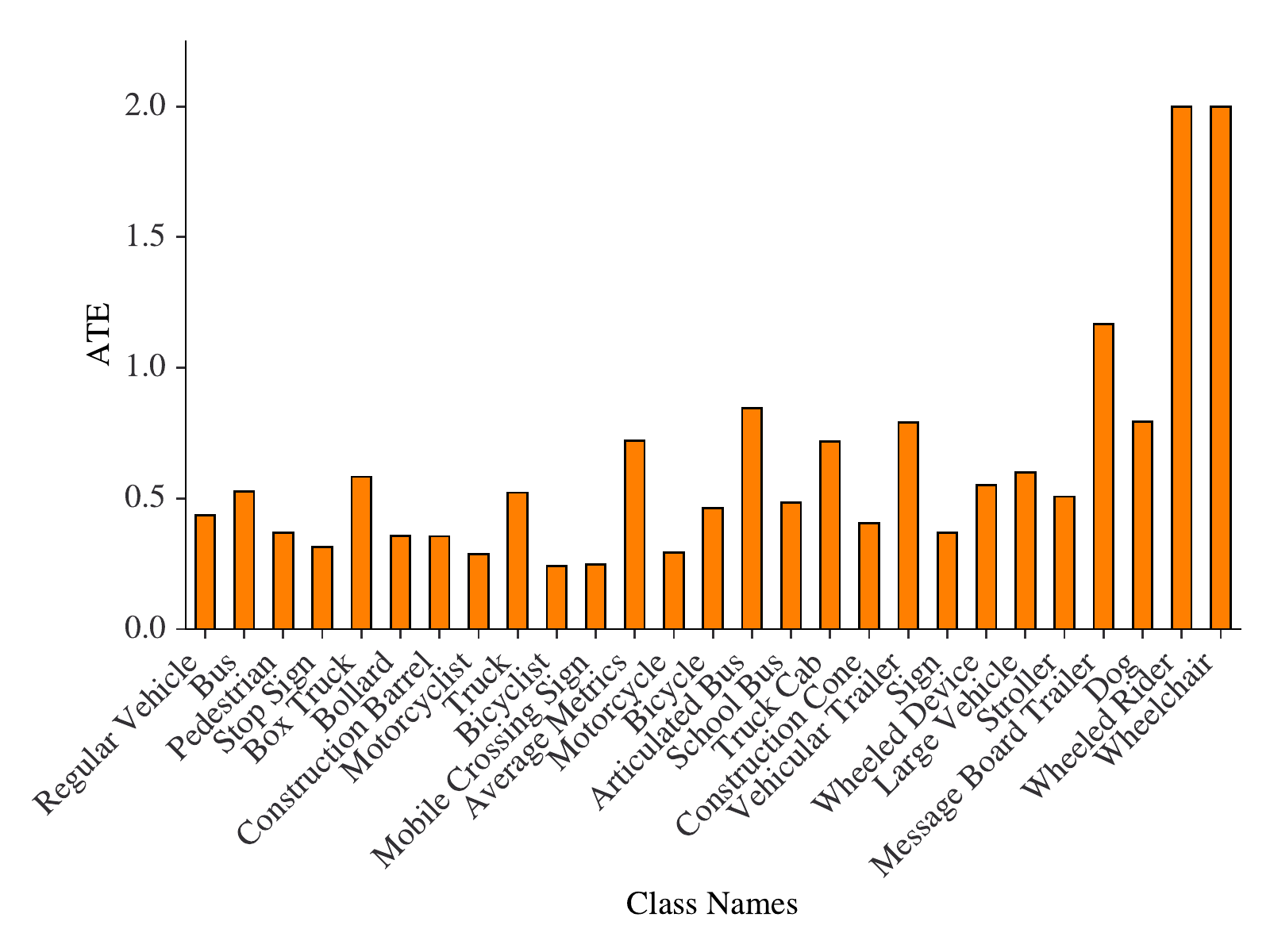}    \includegraphics[scale=0.4]{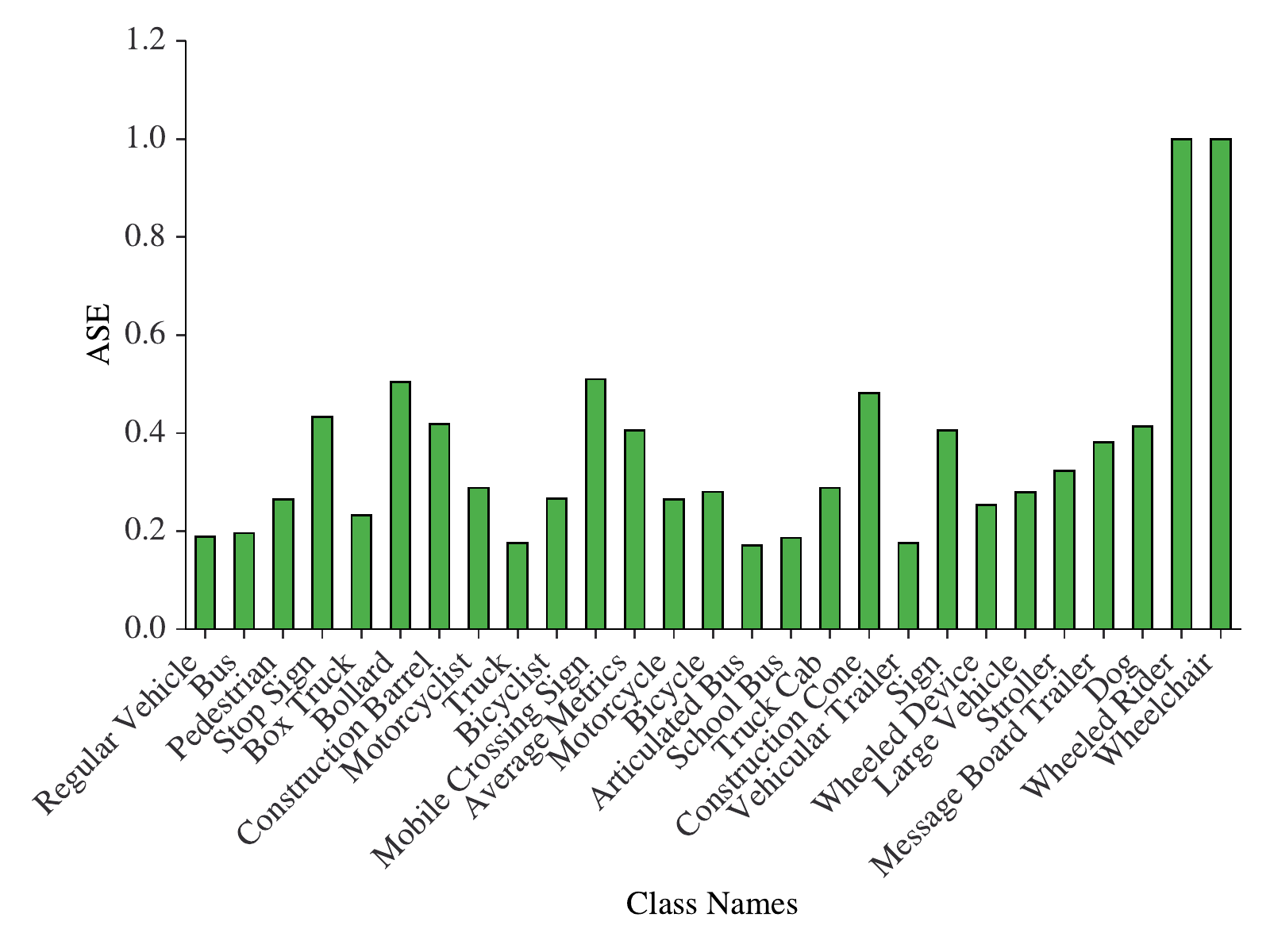}
    \includegraphics[scale=0.4]{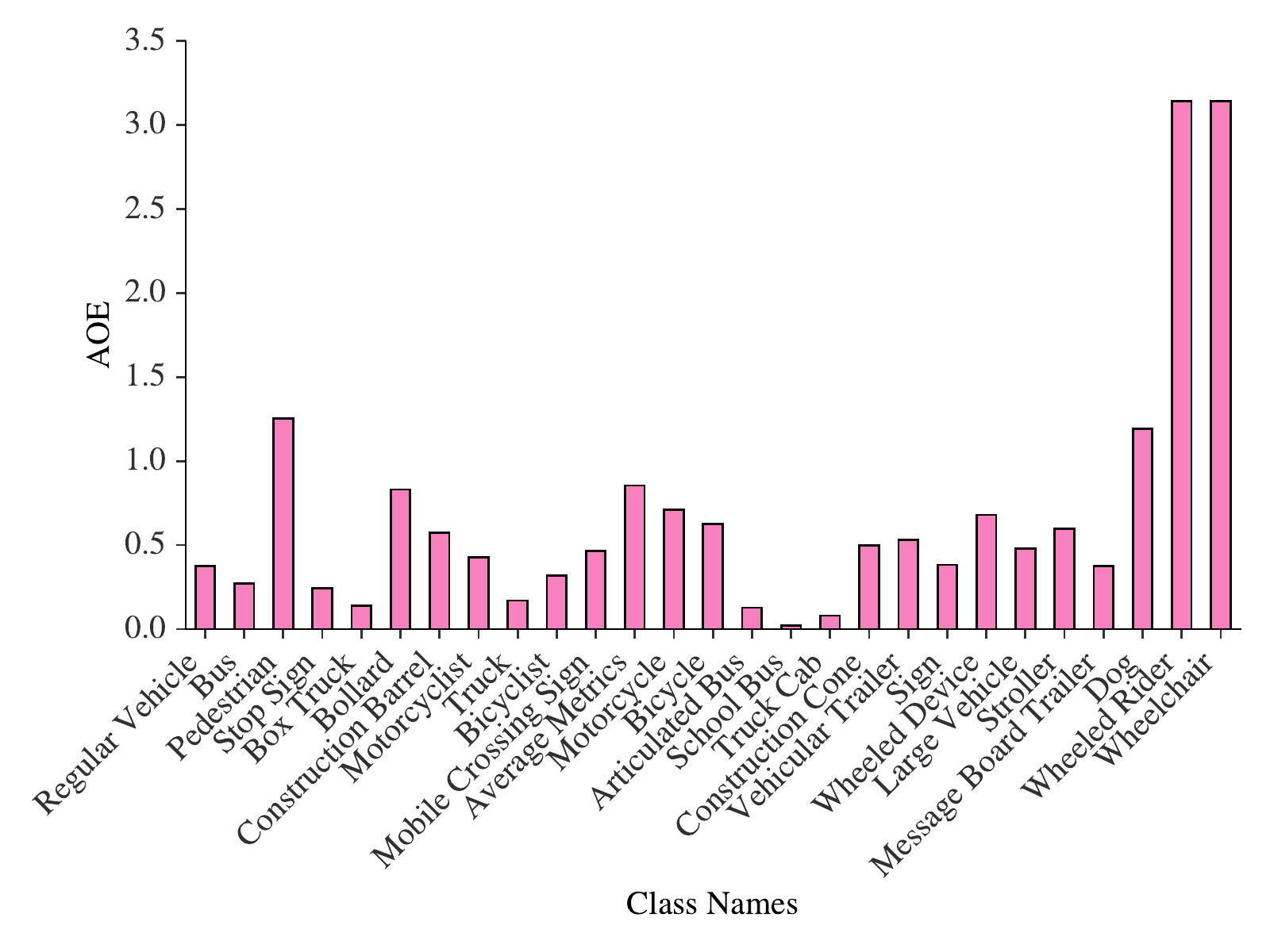}
    \caption{3D object detection performance on the \emph{validation} split of the \emph{Sensor Dataset (Beta)}. \textbf{Top Row:} Composite detection score (left). Average translation error (right) \textbf{Bottom Row:}  Average scaling error (left), and average orientation error (right). Results are shown on the \emph{validation} set of the \emph{Sensor Dataset}.}
    \label{fig:3d-detection-baseline-additional}
\end{figure}

In Figure \ref{fig:3d-detection-baseline-additional}, we show additional evaluation metrics for our detection baseline.

\textbf{Average Precision (AP)}
\begin{equation}
AP = \frac{1}{101} \sum\limits_{t \in \mathcal{T}} \sum\limits_{r \in \mathcal{R}} p_{interp}(r)
\end{equation}

\textbf{True Positive Metrics}
\textbf{Average Translation Error (ATE)}
\begin{equation}
    ATE = \| t_{det} - t_{gt} \|_2
\end{equation}

\textbf{Average Scaling Error (ASE)}
\begin{equation}
    ASE = 1 - \prod\limits_{d \in \mathcal{D}} \frac{\min(d_{det}, d_{gt})}{max(d_{det}, d_{gt})}
\end{equation}

\textbf{Average Orientation Error (AOE)}
\begin{equation}
AOE = | \theta_{det} - \theta_{gt} |
\end{equation}

\textbf{Composite Detection Score (CDS)}
\begin{equation}
CDS = mAP \cdot \sum\limits_{x \in \mathcal{X}} (1 - x)
\end{equation}
where $\mathcal{X} = \{mATE_{unit}, mASE_{unit}, mAOE_{unit} \}$

\pagebreak

\subsection{Training Details of SPF2 baseline}

We sample 2-second training snippets (representing 1 second of past and 1 second of future data) every 0.5 seconds. Thus, for a training log with 30 second duration, 59 training snippets would be sampled. We train the model for 16 epochs by using the Adam optimizer with the learning rate of $4e-3$, betas of 0.9 and 0.999, and batch size of 16 per GPU.

\subsection{Additional Motion Forecasting Experiments} \label{add_mf_experiments}

\subsubsection{Transfer Learning}
The results of transfer learning experiments are summarized in Table \ref{eval-mf-transfer}. WIMP was trained and tested in different settings with Argoverse 1.1 and Argoverse 2. As expected, the model works best when it is trained and tested on the same distribution (i.e. both train and test data come from Argoverse 1.1, or both from Argoverse 2). For example, when WIMP is tested on Argoverse 2 (6s), the model trained on Argoverse 2 (6s) has a \textit{minFDE} of 2.91, whereas the one trained on  Argoverse 1.1 (3s) has a \textit{minFDE} of 6.82 (i.e. approximately 2.3x worse). Likewise, in the reverse setting, when WIMP is tested on Argoverse 1.1 (3s), the model trained on Argoverse 1.1 (3s) has a \textit{minFDE} of 1.14 and the one trained on Argoverse 2 (6s) has \textit{minFDE} of 2.05 (i.e. approximately 1.8x worse). This indicates that transfer learning from \emph{Argoverse 2 (Beta)} to \emph{Argoverse 1.1} is more useful than the reverse setting, despite being smaller in the number of scenarios. However, the publicly released version of  \emph{Argoverse 2 Motion Forecasting} (the non-beta 2.0 version) has comparable size with Argoverse 1.1.

We note that it is a common practice to train and test sequential models on varied sequence length (e.g. machine translation). As such, it is still reasonable to expect a model trained with 3s to do well on 6s horizon.  Several factors may contribute to distribution shift, including differing prediction horizon, cities, mining protocols, object types. Notably, however, these results indicate that Argoverse 2 is significantly more challenging and diverse than its predecessor.

\subsubsection{Experiment with different object types}
Table \ref{eval-mf-object-type} shows the results on Nearest Neighbor baseline (without map prior) on different object types. As one would expect, the displacement errors in pedestrians are significantly lower than other object types. This occurs because they move at significantly slower velocities. However, this does not imply that pedestrian motion forecasting is a solved problem and one should rather focus on other object types. This instead means that we need to come up with better metrics that can capture that fact lower displacement errors in pedestrians can often be more critical than higher errors in vehicles. We leave this line of work for future scope.

\begin{table}[h]
  \caption{Performance of WIMP when trained and tested on different versions of Argoverse motion forecasting datasets.  Training and evaluation is restricted to vehicle-like (vehicle, bus, motorcyclist) object types as only vehicles were present in Argoverse 1.1. All the results are for K=6, and prediction horizon is specified in parentheses.  Notably, the model trained on a 3s horizon performs poorly on the longer 6s horizon. `Argoverse 2' below denotes the \emph{Argoverse 2 (Beta) Motion Forecasting Dataset.}}
  \label{eval-mf-transfer}
  \centering
  \begin{tabular}{c c ccc}
    \toprule
    Train Split (pred. horizon) & Test Split (pred. horizon) & minADE $\downarrow$ & minFDE $\downarrow$ & MR $\downarrow$  \\
    \midrule
    Argoverse 1.1 (3s) & Argoverse 1.1 (3s) & \textbf{0.75} & \textbf{1.14} & \textbf{0.12} \\
    Argoverse 2 (6s) & Argoverse 1.1 (3s) & 1.68 & 2.05 & 0.27 \\
    Argoverse 1.1 (3s) & Argoverse 2 (3s) & 0.94 & 1.88 & 0.26 \\
    \hline
    Argoverse 1.1 (3s) & Argoverse 2 (6s) & 4.93 & 6.82 & 0.77\\
    Argoverse 2 (6s) & Argoverse 2 (6s) & \textbf{1.48} & \textbf{2.91} & \textbf{0.43} \\

    \bottomrule
  \end{tabular}
\end{table}

\begin{table}[ht]
  \caption{Performance of Nearest Neighbor baseline on different object types for K=6. The most accurately predicted object type for each evaluation metric is highlighted in bold.  }
  \label{eval-mf-object-type}
  \centering
  \begin{tabular}{r r ccc}
    \toprule
    Object Type & \#Samples & minADE $\downarrow$ & minFDE $\downarrow$ & MR $\downarrow$  \\
    \midrule
    All & 9955 & 2.48 & 5.52 & 0.66  \\
    Vehicle & 8713 & 2.62 & 5.87 & 0.70 \\
    Bus & 439 & 2.69 & 5.59 & 0.73 \\
    Pedestrians & 677  & \textbf{0.69} & \textbf{1.31} & \textbf{0.17} \\
    Motorcyclist & 39 & 2.33 & 5.07 & 0.61 \\
    Cyclist & 87 & 1.48 & 2.80 & 0.42 \\ 
    \bottomrule
  \end{tabular}
\end{table}

%% file: neurips_data_2021/tables/hd-map-attributes.tex
\begin{table}[htbp]
    \centering
    \begin{adjustbox}{max width=\columnwidth}
    \begingroup
    \renewcommand{\arraystretch}{1.25} %
    \begin{tabular}{c l c l}
    \toprule
    \textsc{\textbf{Map Entity}} & \textsc{\textbf{Provided Attributes}} & \textsc{\textbf{Type}} & \textsc{\textbf{Description}} \\
    \midrule
        \multirow{3}{7em}{\textsc{Lane Segments}} & \textsc{is\_intersection} & \textsc{boolean} & \textsc{whether or not this lane segment lies within an intersection.} \\
                                        & \textsc{lane type} & \textsc{enumerated type} & \textsc{designation of which vehicle types may legally utilize this lane for travel.} \\
                                        & \textsc{left lane boundary} & \textsc{3D polyline} & \textsc{the polyline of the left boundary in the city map coordinate system} \\
                                        & \textsc{right lane boundary} & \textsc{3D polyline} & \textsc{the polyline of the right boundary in the city map coordinate system.} \\
                                        & \textsc{left lane mark type} & \textsc{enumerated type} & \textsc{type of painted lane marking to the left of the lane segment on the road.}\\ %
                                        & \textsc{right lane mark type} & \textsc{enumerated type} & \textsc{type of painted lane marking to the right of the lane segment on the road.}\\ %
                                        & \textsc{left neighbor} & \textsc{integer} & \textsc{the unique lane segment immediately to the left of segment, or none.} \\
                                        & \textsc{right neighbor} & \textsc{integer} & \textsc{the unique lane segment immediately to the right of segment, or none.} \\
                                        & \textsc{successor IDs} & \textsc{integer list} & \textsc{lane segments that may be entered by following forward. } \\
                                        & \textsc{ID} & \textsc{integer} & \textsc{unique identifier} \\
        \hline
        \multirow{2}{7em}{\textsc{Drivable Area}} & \textsc{area boundary} & \textsc{3D polygons} & \textsc{area where it is possible for the AV to drive without damaging itself} \\
                                         & \textsc{ID} & \textsc{integer} & \textsc{unique identifier} \\
    \hline
        \multirow{2}{7em}{\textsc{Pedestrian Crossings}} & \textsc{Edge1, Edge2} & \textsc{3D polylines} & \textsc{endpoints of both edge along the principal axis, thus defining a polygon.} \\
                                         & \textsc{ID} & \textsc{integer} & \textsc{unique identifier} \\
    \hline
        \textsc{Ground surface height} & - & \textsc{2d raster array} & \textsc{Raster grid quantized to a \SI{30}{\centi\meter} resolution.} \\
    \bottomrule
    \end{tabular}
    \endgroup
    \end{adjustbox}
    \vspace{0.5em}
    \caption{HD map attributes for each Argoverse 2 scenario.}
    \label{tab:mapentities}
\end{table}

%% file: neurips_data_2021/tables/sensor-dataset-map-examples-with-crosswalk.tex
\begin{figure}[htbp]
    \centering

    \subfloat[Washington D.C.]{
    	\includegraphics[width=0.3\linewidth]{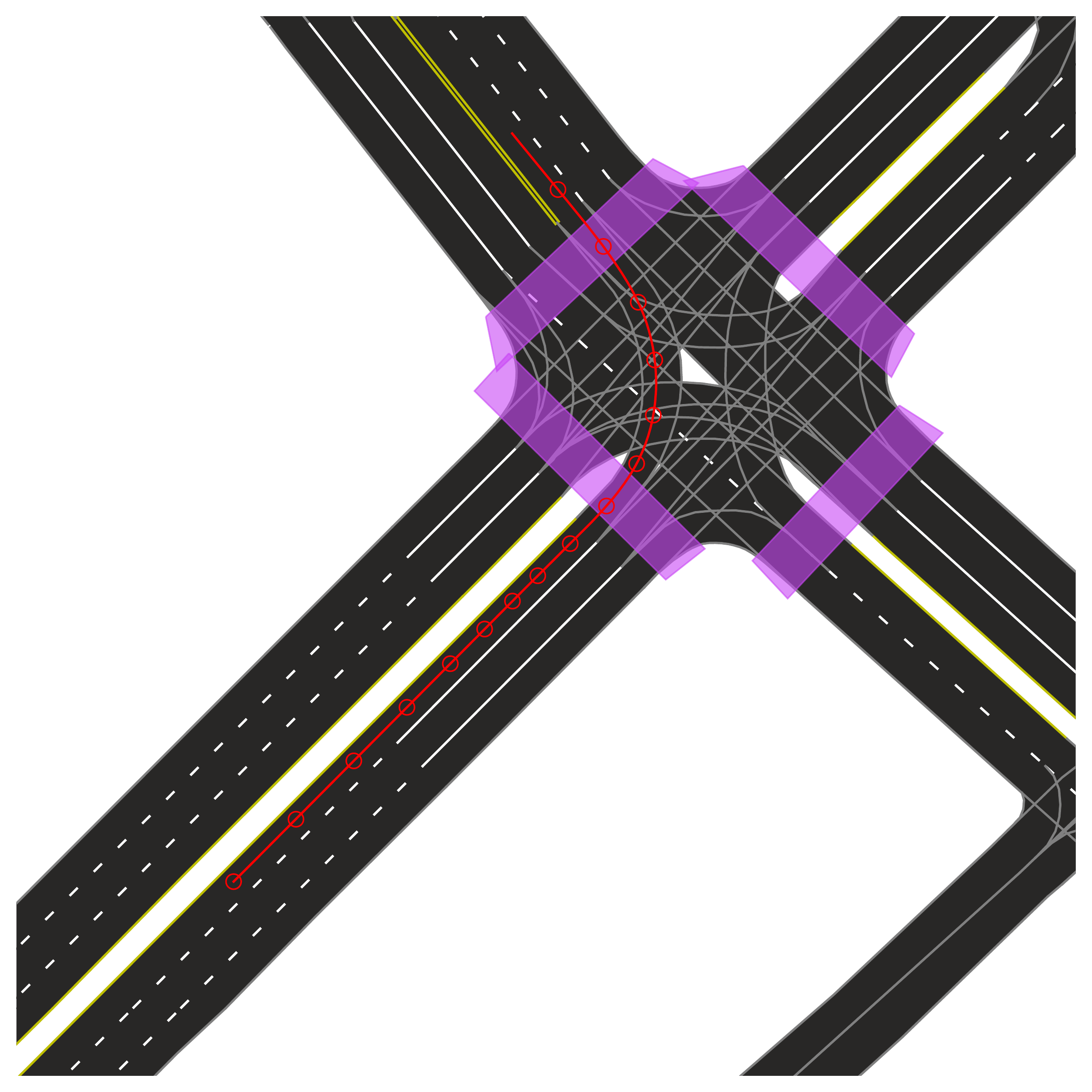}
    }
    \subfloat[Washington D.C.]{
    	\includegraphics[width=0.3\linewidth]{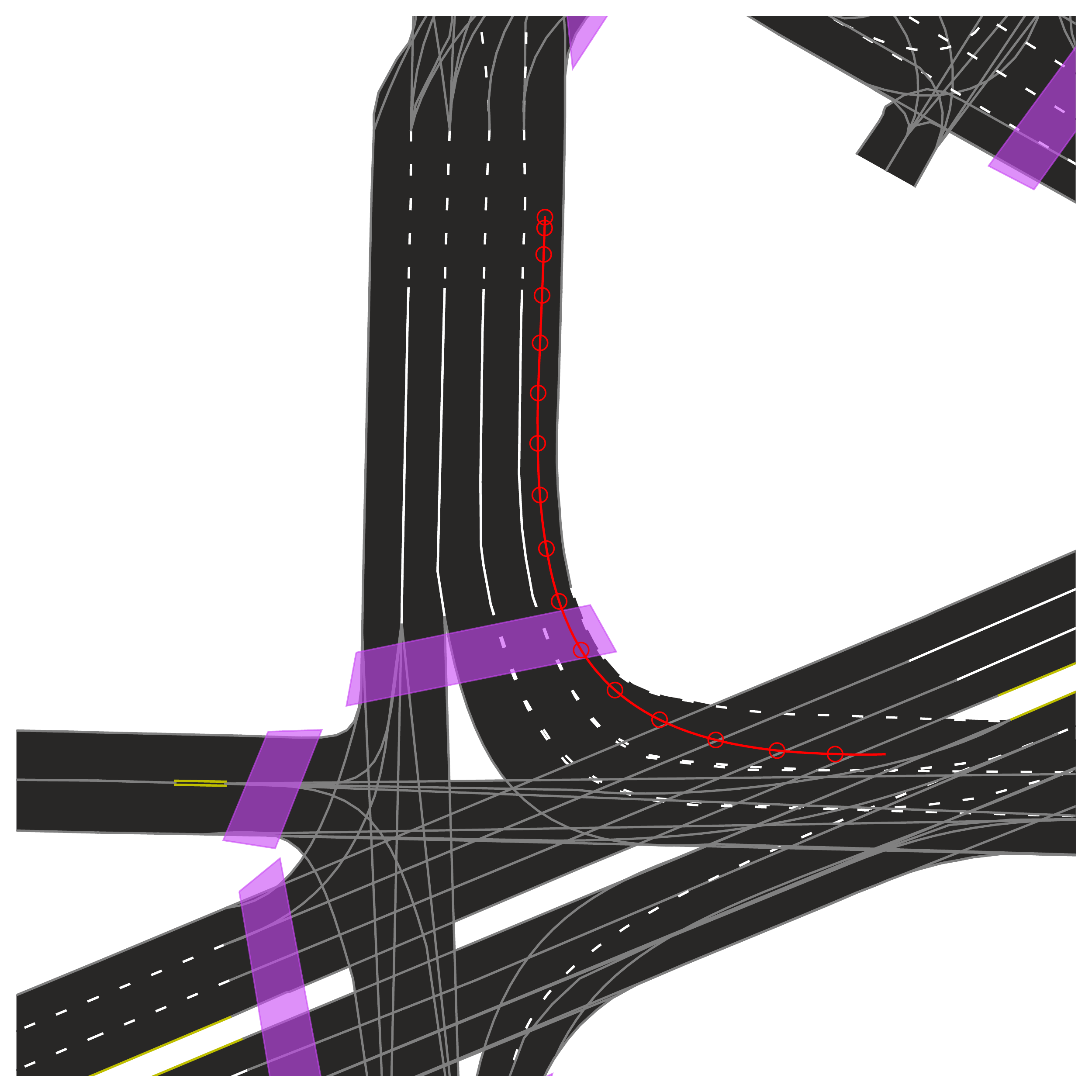}
    }
    \subfloat[Washington D.C.]{
    	\includegraphics[width=0.3\linewidth]{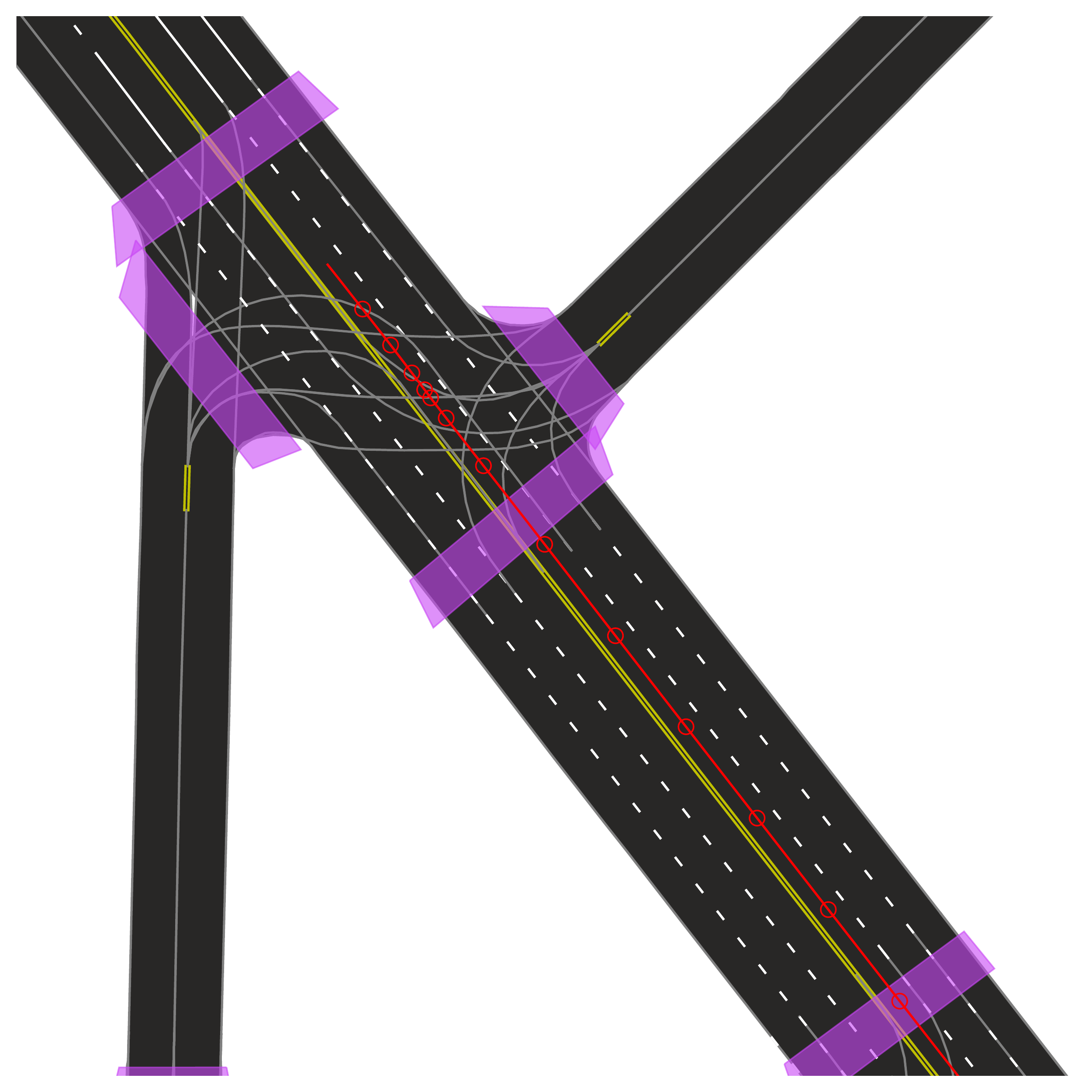}
    }
    \hspace{0mm}
    \subfloat[Pittsburgh, PA]{
    	\includegraphics[width=0.3\linewidth]{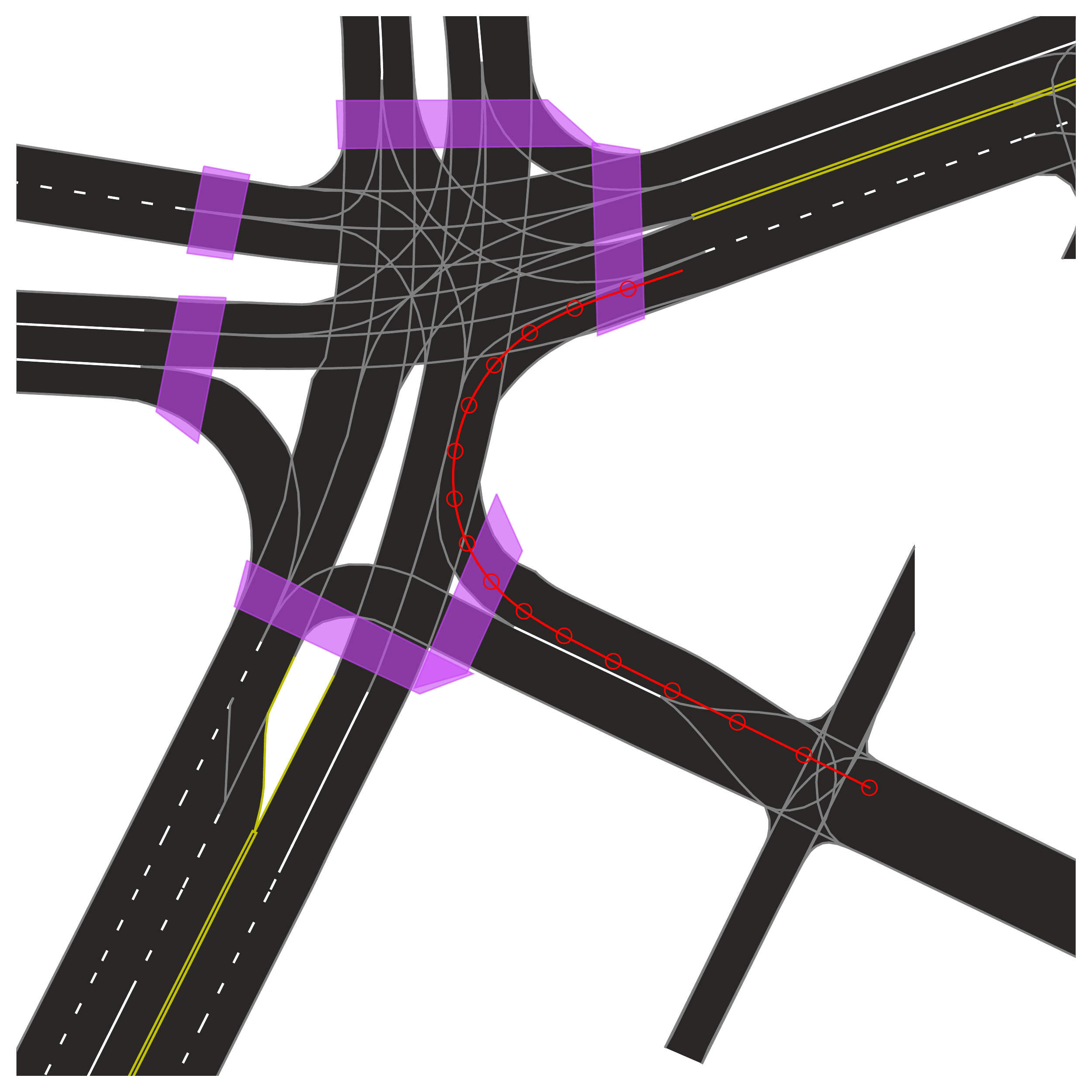}
    }
    \subfloat[Pittsburgh, PA]{
    	\includegraphics[width=0.3\linewidth]{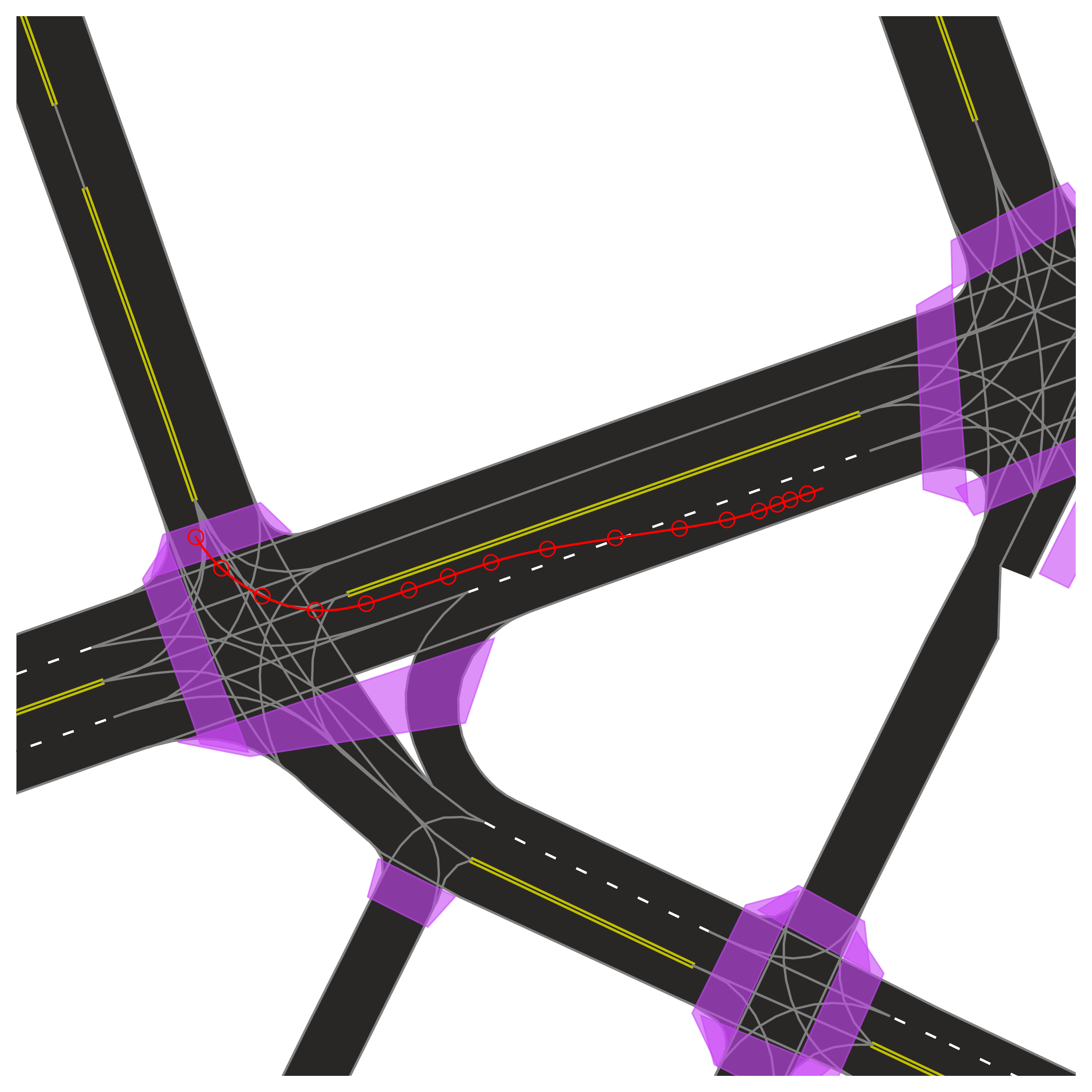}
    }
    \subfloat[Pittsburgh, PA]{
    	\includegraphics[width=0.3\linewidth]{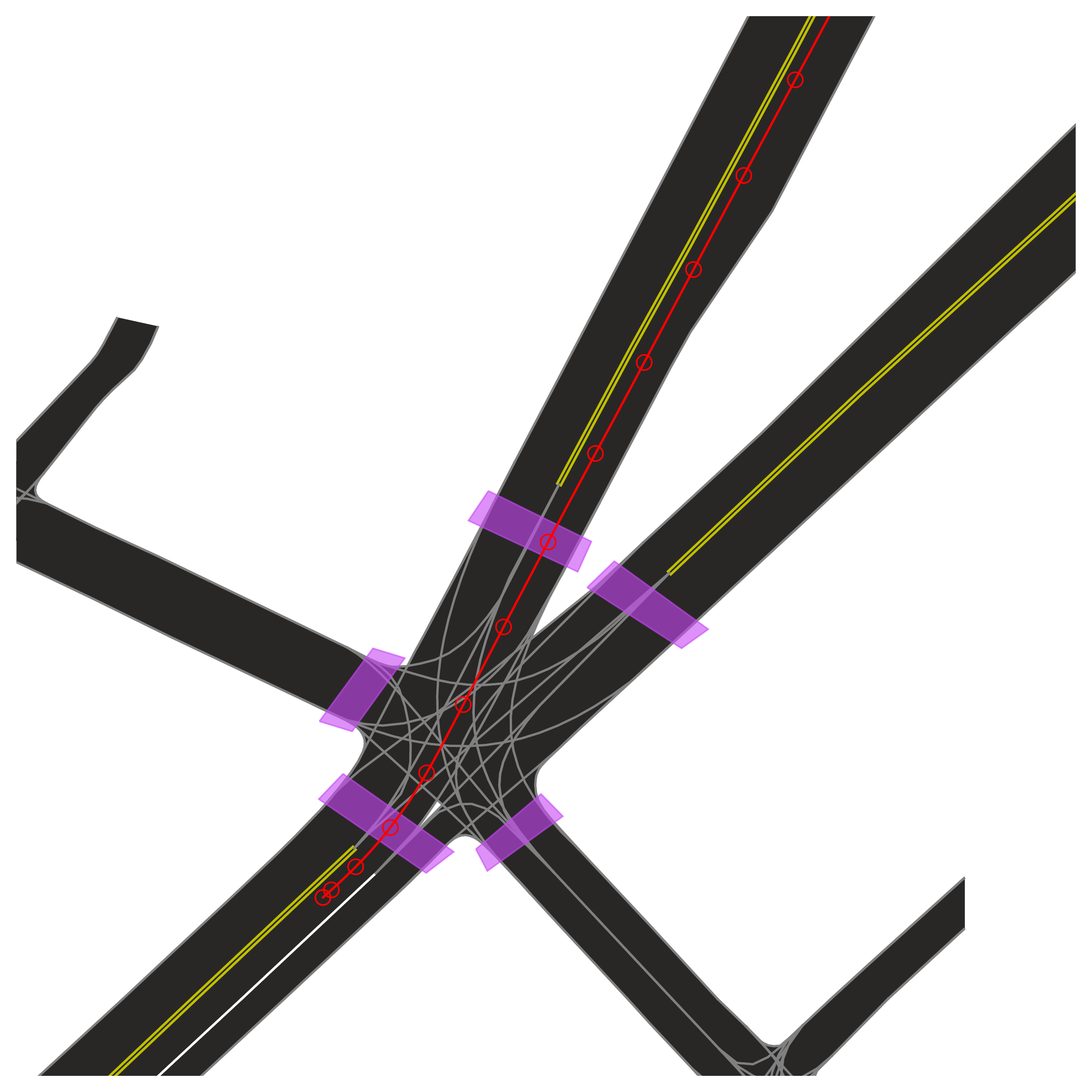}
    }
    \hspace{0mm}
    \subfloat[Miami, FL]{
    	\includegraphics[width=0.3\linewidth]{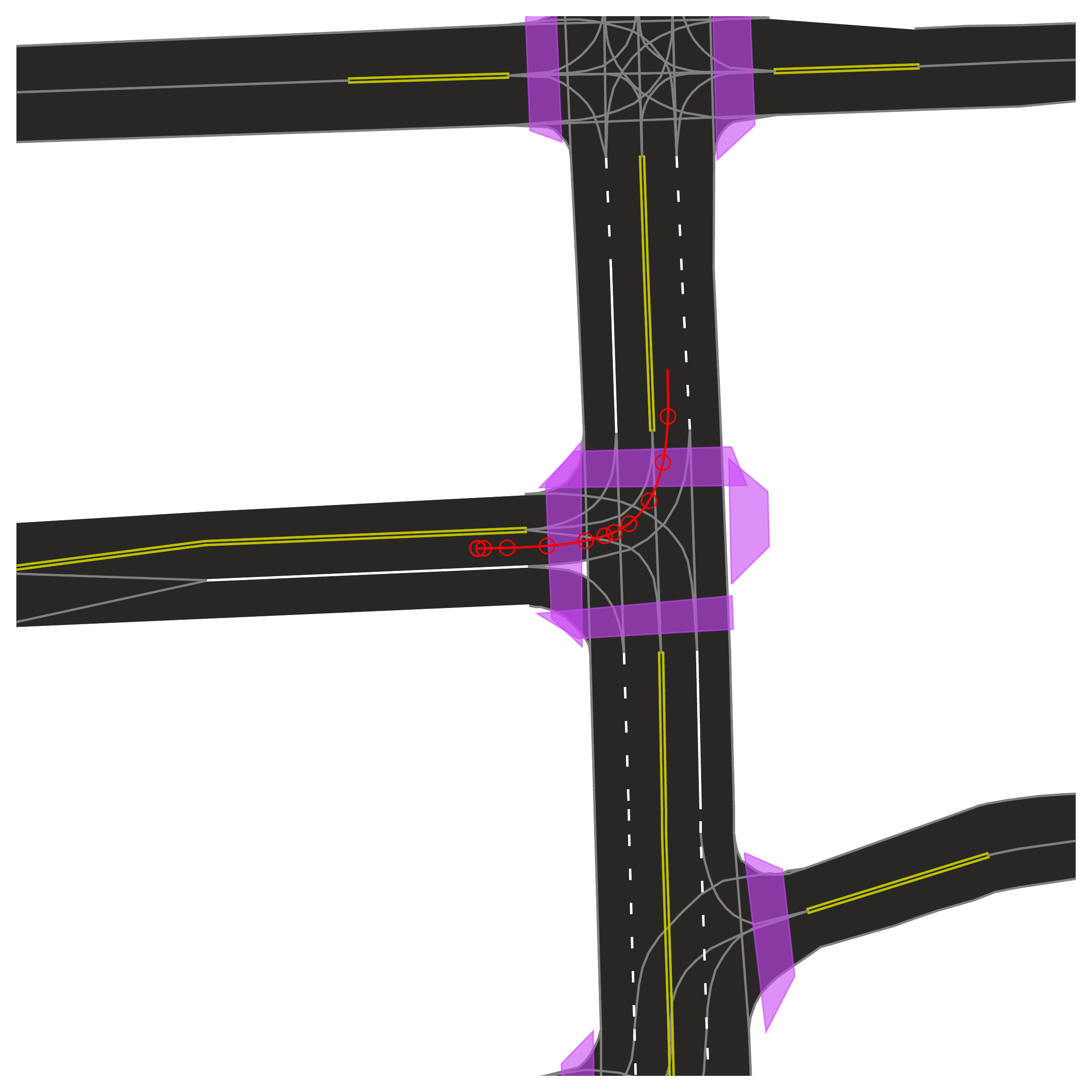}
    }
    \subfloat[Miami, FL]{
    	\includegraphics[width=0.3\linewidth]{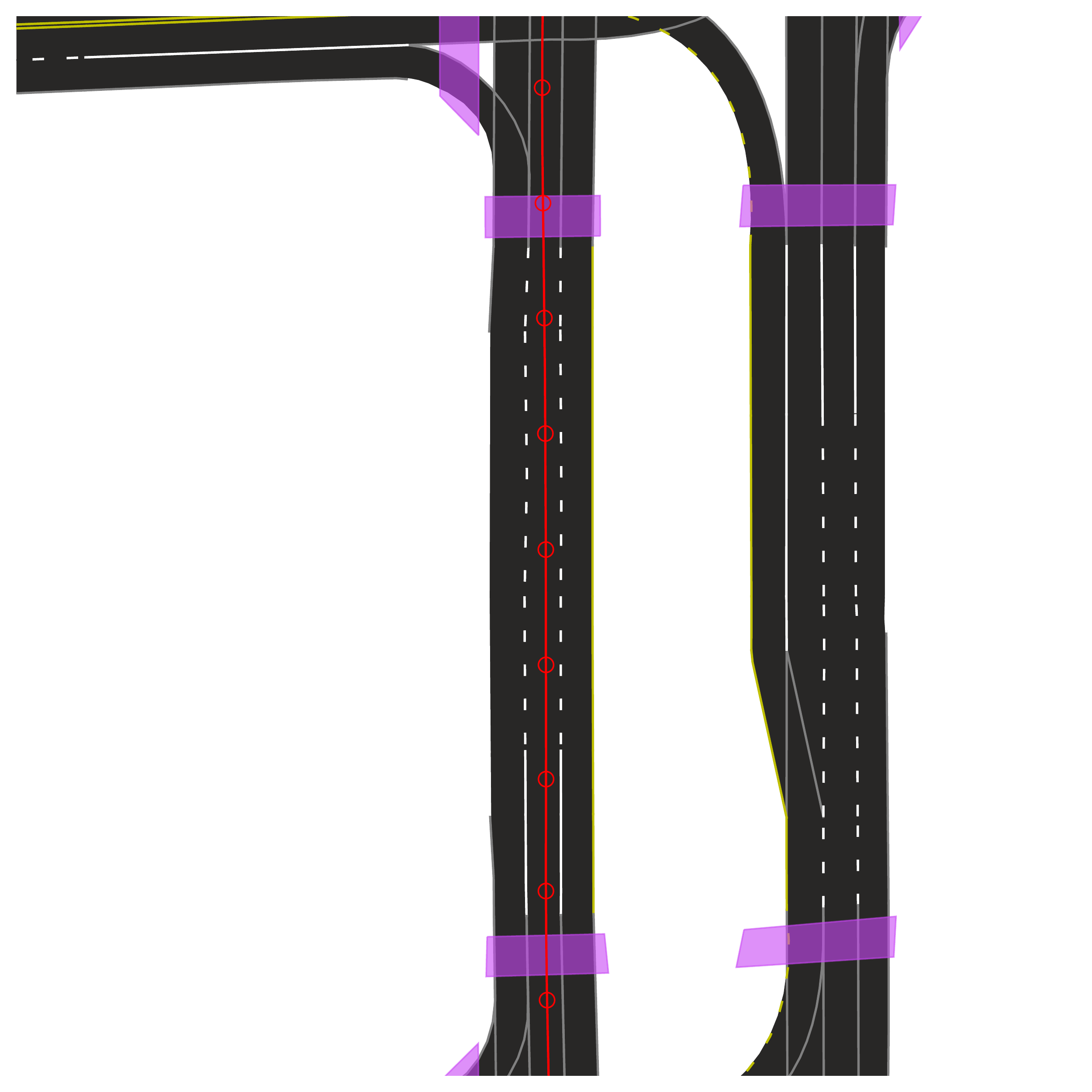}
    }
    \subfloat[Miami, FL]{
    	\includegraphics[width=0.3\linewidth]{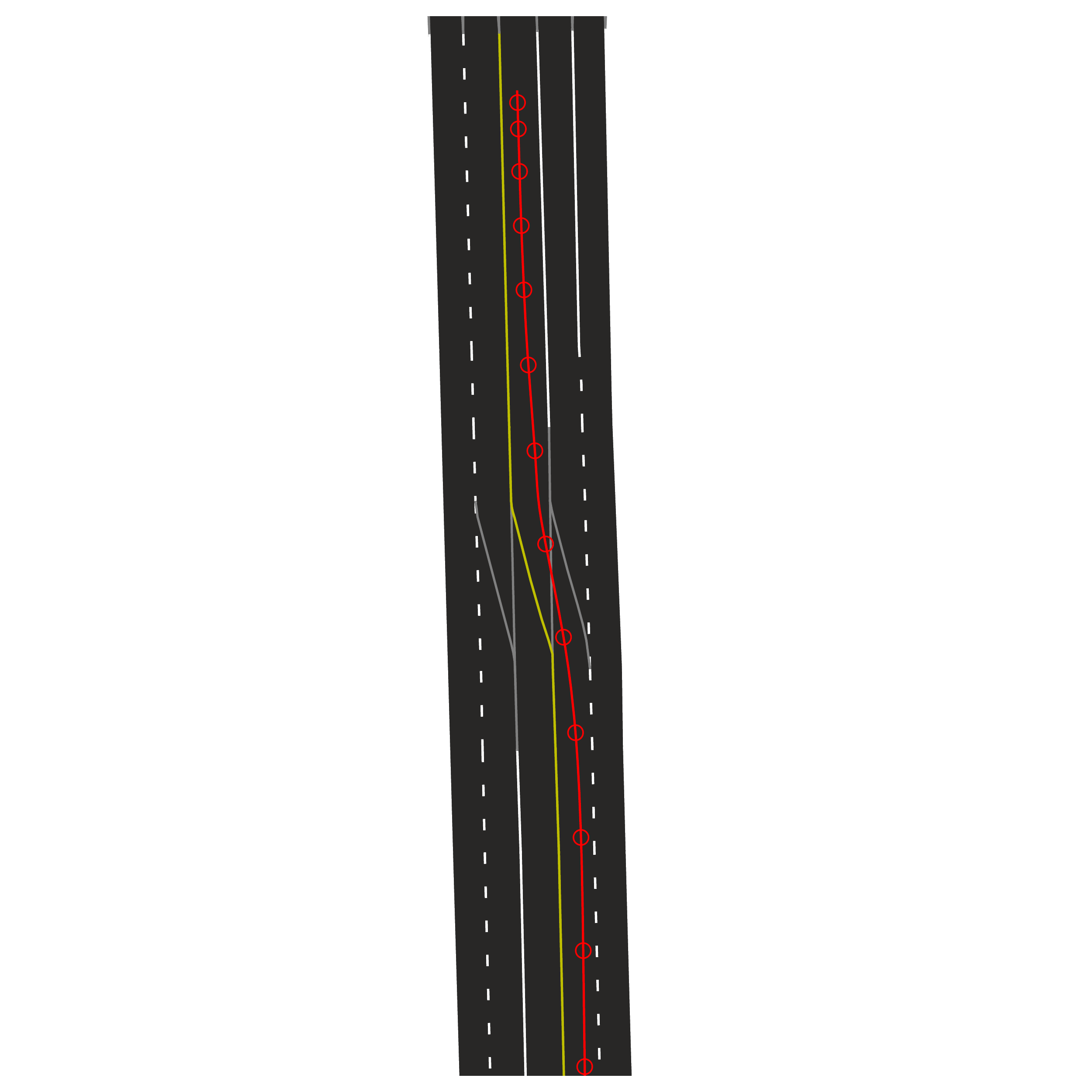}
    }
    \hspace{0mm}
    \subfloat[Detroit, MI]{
    	\includegraphics[width=0.3\linewidth]{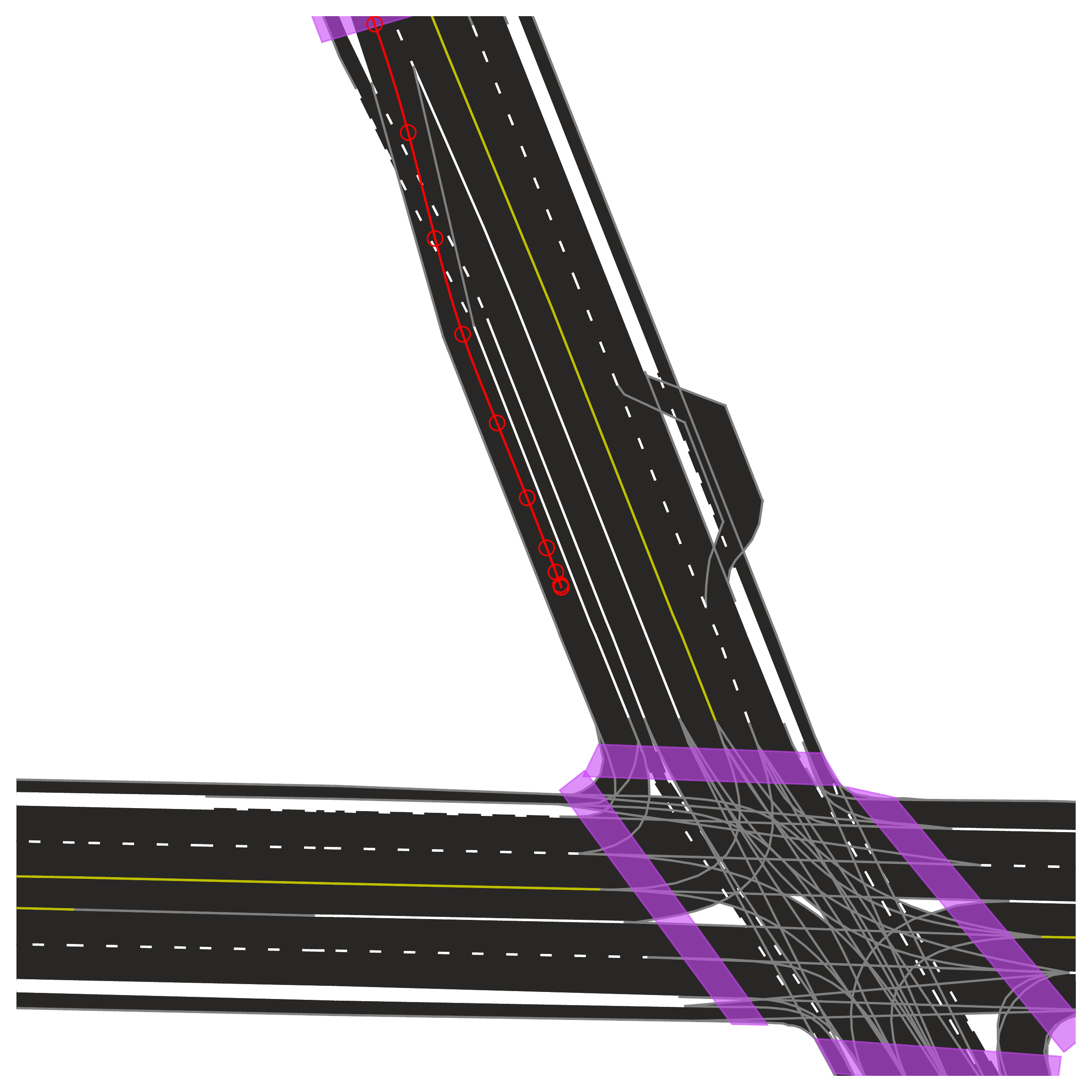}
    }
    \subfloat[Detroit, MI]{
    	\includegraphics[width=0.3\linewidth]{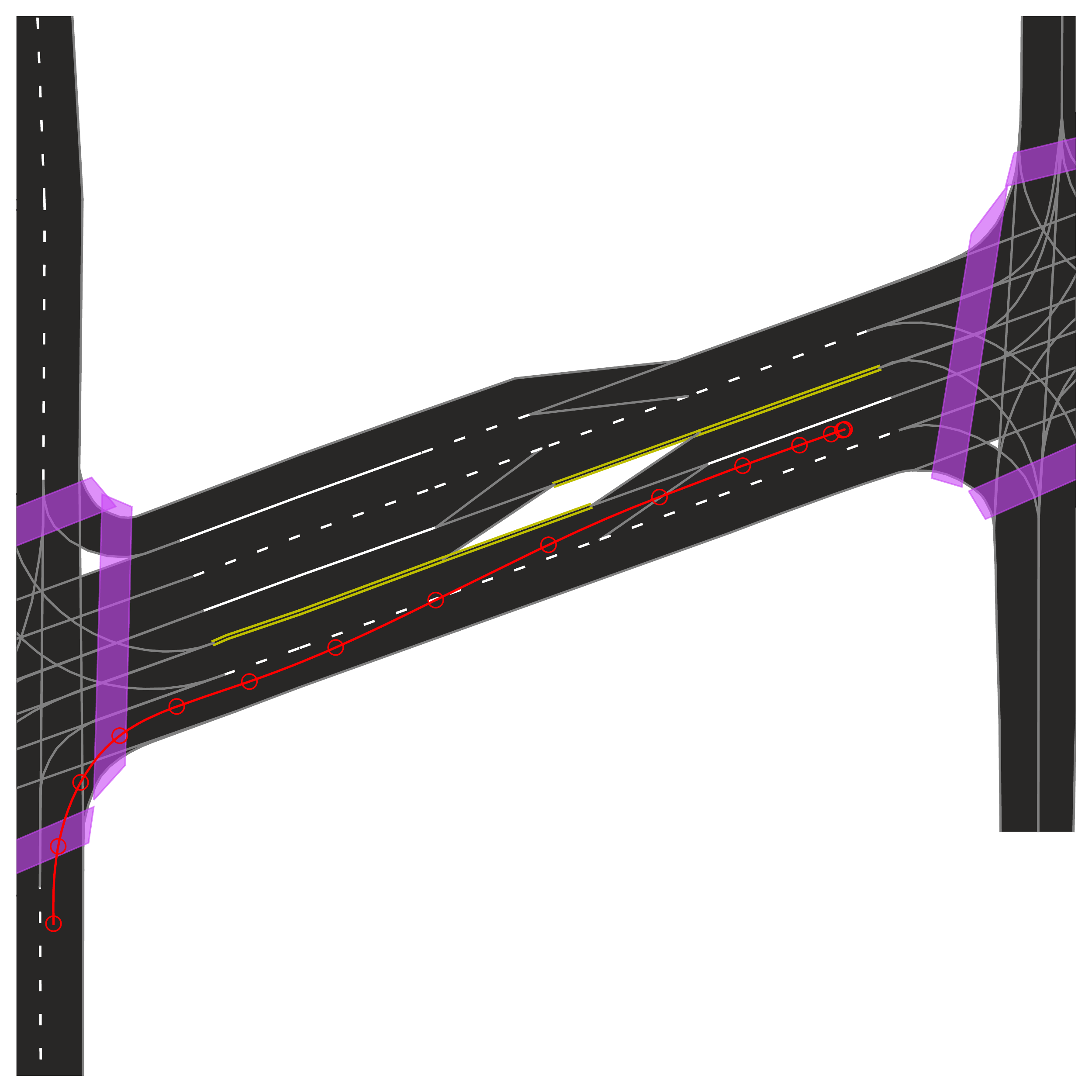}
    }
    \subfloat[Austin, TX]{
    	\includegraphics[width=0.3\linewidth]{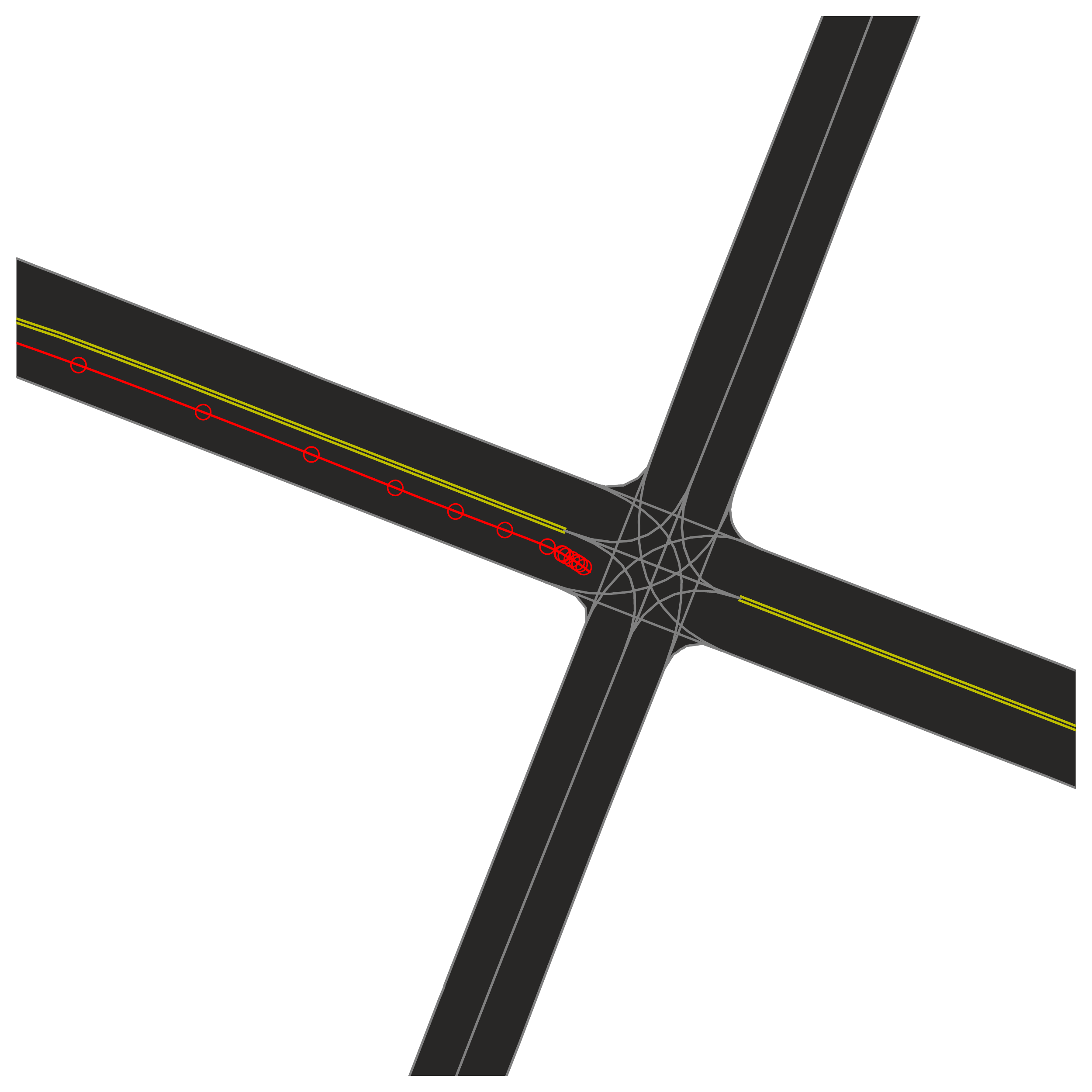}
    }
    \caption{Examples of egovehicle (AV) trajectories on local vector maps from the Sensor Dataset across several different cities. A 100m $\times$ 100m local map region is shown. Crosswalks are indicated in purple. Red circles denote the AV pose discretely sampled at 1 Hz for the purposes of illustration.  Pose is provided at >20 Hz in the dataset, as indicated by the trajectory path indicated by a red line. City layouts vary dramatically, e.g. roadways in Miami are usually aligned parallel to a north-south, east-west grid, while roadways in Pittsburgh are generally not. }
    \label{fig:sensordatasetmapexamples}
\end{figure}

%% file: neurips_data_2021/sections/datasheet.tex
\section{Datasheet for Argoverse 2}
\textcolor{\sectioncolor}{\textbf{
    For what purpose was the dataset created?
    }
    Was there a specific task in mind? Was there
    a specific gap that needed to be filled? Please provide a description.
    } \\
    Argoverse was created to support the global research community in improving the state of the art in machine learning tasks vital for self driving. The Argoverse 2 datasets described in this manuscript improve upon the initial Argoverse datasets. These datasets support many tasks, from 3D perception to motion forecasting to HD map automation.
    
    The three datasets proposed in this manuscript address different gaps in this space. See the comparison charts in the main manuscript for a more detailed breakdown. 
    
    The Argoverse 2 \emph{Sensor Dataset} has a richer taxonomy than similar datasets. It is the only dataset of similar size to have stereo imagery. The 1,000 logs in the dataset were chosen to have a variety of object types with diverse interactions.
    
    The Argoverse 2 \emph{Motion Forecasting Dataset} also has a richer taxonomy than existing datasets. The scenarios in the dataset were mined with an emphasis on unusual behaviors that are difficult to predict.
    
    The Argoverse 2 \emph{Lidar Dataset} is the largest \emph{Lidar Dataset}. Only the concurrent ONCE dataset is similarly sized to enable self-supervised learning in lidar space. Unlike ONCE, our dataset contains HD maps and high frame rate lidar.

    \textcolor{\sectioncolor}{\textbf{
    Who created this dataset (e.g., which team, research group) and on behalf
    of which entity (e.g., company, institution, organization)?
    }
    } \\
    The Argoverse 2 datasets were created by researchers at Argo AI. \\
    
    \textcolor{\sectioncolor}{\textbf{
    What support was needed to make this dataset?
    }
    (e.g.who funded the creation of the dataset? If there is an associated
    grant, provide the name of the grantor and the grant name and number, or if
    it was supported by a company or government agency, give those details.)
    } \\
    The creation of this dataset was funded by Argo AI. \\
    
    \textcolor{\sectioncolor}{\textbf{
    Any other comments?
    }} \\
    n/a \\

\begin{mdframed}[linecolor=\sectioncolor]
\section*{\textcolor{\sectioncolor}{
    COMPOSITION
}}
\end{mdframed}
    \textcolor{\sectioncolor}{\textbf{
    What do the instances that comprise the dataset represent (e.g., documents,
    photos, people, countries)?
    }
    Are there multiple types of instances (e.g., movies, users, and ratings;
    people and interactions between them; nodes and edges)? Please provide a
    description.
    } \\
    
    The three constituent datasets of Argoverse 2 have different attributes, but the core instances for each are brief ``scenarios'' or ``logs'' of 11, 15, or 30 seconds that represent a continuous observation of a scene around a self-driving vehicle. 
    
    Each scenario in all three datasets has an HD map that includes lane boundaries, crosswalks, driveable area, etc. Scenarios for the \emph{Sensor Dataset} additionally contain a raster map of ground height at .3 meter resolution.
    
    \textcolor{\sectioncolor}{\textbf{
    How many instances are there in total (of each type, if appropriate)?
    }
    } \\
    The \emph{Sensor Dataset} has 1,000 15 second scenarios. 
    
    The \emph{Lidar Dataset} has 20,000 30 second scenarios. 
    
    The \emph{Motion Forecasting Dataset} has 250,000 11 second scenarios. 
    
    \textcolor{\sectioncolor}{\textbf{
    Does the dataset contain all possible instances or is it a sample (not
    necessarily random) of instances from a larger set?
    }
    If the dataset is a sample, then what is the larger set? Is the sample
    representative of the larger set (e.g., geographic coverage)? If so, please
    describe how this representativeness was validated/verified. If it is not
    representative of the larger set, please describe why not (e.g., to cover a
    more diverse range of instances, because instances were withheld or
    unavailable).
    } \\
    The scenarios in the dataset are a sample of the set of observations made by a fleet of self-driving vehicles. The data is not uniformly sampled. The particular samples were chosen to be geographically diverse (spanning 6 cities - Pittsburgh, Detroit, Austin, Palo Alto, Miami, and Washington D.C.), to include interesting behavior (e.g. cars making unexpected maneuvers), to contain interesting weather (e.g. rain and snow), and to contain scenes with many objects of diverse types in motion (e.g. a crowd walking, riders on e-scooters splitting lanes between many vehicles, an excavator operating at a construction site, etc.).
    
    \textcolor{\sectioncolor}{\textbf{
    What data does each instance consist of?
    }
    “Raw” data (e.g., unprocessed text or images) or features? In either case,
    please provide a description.
    } \\
    Each \emph{Sensor Dataset} scenario is 15 seconds in duration. Each scenario has 20 fps video from 7 ring cameras, 20 fps video from two forward facing stereo cameras, and 10 hz lidar returns from two out-of-phase 32 beam lidars. The ring cameras are synchronized to fire when either lidar sweeps through their field of view. Each scenario contains vehicle pose over time and calibration data to relate the various sensors.
    
    Each \emph{Lidar Dataset} scenario is 30 seconds in duration. These scenarios are similar to those of the \emph{Sensor Dataset}, except that there is no imagery. 
    
    Each \emph{Motion Forecasting} scenario is 11 seconds in duration. These scenarios contain no sensor data, but instead contain tracks of objects such as vehicles, pedestrians, and bicycles. The tracks specify the category of each object (e.g. bus or bicycle) as well as their location and heading at a 10 hz sampling interval.
    
    The HD map associated with all three types of scenarios contains polylines describing lanes, crosswalks, and driveable area. Lanes form a graph with predecessors and successors, e.g. a lane that splits can have two successors. Lanes have precisely localized lane boundaries that include paint type (e.g. double solid yellow). Driveable area, also described by a polygon, is the area where it is possible but not necessarily legal to drive. It includes areas such as road shoulders.
    
    \textcolor{\sectioncolor}{\textbf{
    Is there a label or target associated with each instance?
    }
    If so, please provide a description.
    } \\
    Each \emph{Sensor Dataset} scenario has 3D track annotations for dynamic objects such as vehicles, pedestrians, strollers, dogs, etc. The tracks are suitable as ground truth for tasks such as 3D object detection and 3D tracking. The 3D track labels are intentionally held out from the test set. The HD map could also be thought of as labels for each instance, and would be suitable as ground truth for lane detection or map automation. The vehicle pose data could be considered ground truth labels for visual odometry. The lidar depth estimates can act as ground truth for monocular or stereo depth estimation.
    
    The \emph{Lidar Dataset} does not have human annotations beyond the HD map. Still, the evolving point cloud itself can be considered ground truth for point cloud forecasting.
    
    Each \emph{Motion Forecasting Dataset} scenario provides labels specifying which tracks are associated with ``scored actors''. These tracks exhibit interesting behavior and are guaranteed to be observed over the entire duration of each scenario; algorithms will be asked to forecast the future motion for these tracks. The future motion of actors in each scenario is intentionally held out in the test set.
    
    \textcolor{\sectioncolor}{\textbf{
    Is any information missing from individual instances?
    }
    If so, please provide a description, explaining why this information is
    missing (e.g., because it was unavailable). This does not include
    intentionally removed information, but might include, e.g., redacted text.
    } \\
    In the \emph{Sensor Dataset}, objects are only labeled within 5 meters of the driveable area. For example, a person sitting on their front porch will not be labeled. \\
    
    In the \emph{Sensor Dataset} and \emph{Motion Forecasting Dataset}, instances are not necessarily labeled for the full duration of each scenario if the objects move out of observation range or become occluded. 
    
    \textcolor{\sectioncolor}{\textbf{Z
    Are relationships between individual instances made explicit (e.g., users’
    movie ratings, social network links)?
    }
    If so, please describe how these relationships are made explicit.
    } \\
    The instances of the three datasets are disjoint. They each carry their own HD map for the region around the scenario. These HD maps may overlap spatially, though. For example, many forecasting scenarios may take place in the same intersection. If a user of the dataset wanted to recover the spatial relationship between scenarios, they could do so through the Argoverse API. \\
    
    \textcolor{\sectioncolor}{\textbf{
    Are there recommended data splits (e.g., training, development/validation,
    testing)?
    }
    If so, please provide a description of these splits, explaining the
    rationale behind them.
    } \\
    We define splits of each dataset. The \emph{Sensor Dataset} is split 700 / 150 / 150 between train, validation, and test. The \emph{Lidar Dataset} is split 16,000 / 2,000 / 2,000 and the \emph{Motion Forecasting Dataset} is split 200,000 / 25,000 / 25,000. In all cases, the splits are designed to make the training dataset as large as possible while keeping the validation and test datasets large and diverse enough to accurately benchmark models learned on the training set. \\
    
    \textcolor{\sectioncolor}{\textbf{
    Are there any errors, sources of noise, or redundancies in the dataset?
    }
    If so, please provide a description.
    } \\
    Every sensor used in the dataset -- ring cameras, stereo cameras, and lidar -- has noise associated with it. Pixel intensities, lidar intensities, and lidar point 3D locations all have noise. Lidar points are also quantized to float16 which leads to roughly a centimeter of quantization error. Six degree of freedom vehicle pose also has noise. The calibration specifying the relationship between sensors can be imperfect.
    
    The HD map for each scenario can contain noise, both in terms of lane boundary locations and precise ground height.
    
    The 3D object annotations in the \emph{Sensor Dataset} do not always match the spatial extent and motion of an object in the real world. For example, we assume that objects do not change size during a scenario, but this could be violated by a car opening its door. 3D annotations for distant objects with relatively few pixels and lidar returns are less accurate.
    
    The object tracks in the \emph{Motion Forecasting} dataset are imperfect and contain errors typical of a real-time 3D tracking method. Our expectation is that a motion forecasting algorithm should operate well despite this noise.
    
    \textcolor{\sectioncolor}{\textbf{
    Is the dataset self-contained, or does it link to or otherwise rely on
    external resources (e.g., websites, tweets, other datasets)?
    }
    If it links to or relies on external resources, a) are there guarantees
    that they will exist, and remain constant, over time; b) are there official
    archival versions of the complete dataset (i.e., including the external
    resources as they existed at the time the dataset was created); c) are
    there any restrictions (e.g., licenses, fees) associated with any of the
    external resources that might apply to a future user? Please provide
    descriptions of all external resources and any restrictions associated with
    them, as well as links or other access points, as appropriate.
    } \\
    The data itself is self-hosted, like Argoverse 1 [see \url{https://www.argoverse.org/}], and we maintain public links to all previous versions of the dataset in case of updates. The data is independent of any previous datasets, including Argoverse 1.

    \textcolor{\sectioncolor}{\textbf{
    Does the dataset contain data that might be considered confidential (e.g.,
    data that is protected by legal privilege or by doctor-patient
    confidentiality, data that includes the content of individuals’ non-public
    communications)?
    }
    If so, please provide a description.
    } \\
    No. \\
    
    \textcolor{\sectioncolor}{\textbf{
    Does the dataset contain data that, if viewed directly, might be offensive,
    insulting, threatening, or might otherwise cause anxiety?
    }
    If so, please describe why.
    } \\
    No.\\
    
    \textcolor{\sectioncolor}{\textbf{
    Does the dataset relate to people?
    }
    If not, you may skip the remaining questions in this section.
    } \\
    Yes, the dataset contains images and behaviors of thousands of people on public streets. \\
    
    \textcolor{\sectioncolor}{\textbf{
    Does the dataset identify any subpopulations (e.g., by age, gender)?
    }
    If so, please describe how these subpopulations are identified and
    provide a description of their respective distributions within the dataset.
    } \\
    No. \\
    
    \textcolor{\sectioncolor}{\textbf{
    Is it possible to identify individuals (i.e., one or more natural persons),
    either directly or indirectly (i.e., in combination with other data) from
    the dataset?
    }
    If so, please describe how.
    } \\
    We do not believe so. All image data has been anonymized. Faces and license plates are obfuscated by replacing them with a 5x5 grid, where each grid cell is the average color of the original pixels in that grid cell. This anonymization is done manually and is not limited by our 3D annotation policy. For example, a person sitting on their front porch 10 meters from the road would not be labeled with a 3D cuboid, but their face would still be obscured.\\
    
    \textcolor{\sectioncolor}{\textbf{
    Does the dataset contain data that might be considered sensitive in any way
    (e.g., data that reveals racial or ethnic origins, sexual orientations,
    religious beliefs, political opinions or union memberships, or locations;
    financial or health data; biometric or genetic data; forms of government
    identification, such as social security numbers; criminal history)?
    }
    If so, please provide a description.
    } \\
    No. \\
    
    \textcolor{\sectioncolor}{\textbf{
    Any other comments?
    }} \\
    n/a \\

\begin{mdframed}[linecolor=\sectioncolor]
\section*{\textcolor{\sectioncolor}{
    COLLECTION
}}
\end{mdframed}

    \textcolor{\sectioncolor}{\textbf{
    How was the data associated with each instance acquired?
    }
    Was the data directly observable (e.g., raw text, movie ratings),
    reported by subjects (e.g., survey responses), or indirectly
    inferred/derived from other data (e.g., part-of-speech tags, model-based
    guesses for age or language)? If data was reported by subjects or
    indirectly inferred/derived from other data, was the data
    validated/verified? If so, please describe how.
    } \\
    The sensor data was directly acquired by a fleet of autonomous vehicles.\\
    
    \textcolor{\sectioncolor}{\textbf{
    Over what timeframe was the data collected?
    }
    Does this timeframe match the creation timeframe of the data associated
    with the instances (e.g., recent crawl of old news articles)? If not,
    please describe the timeframe in which the data associated with the
    instances was created. Finally, list when the dataset was first published.
    } \\
    The data was collected in 2020 and 2021. The dataset was made public after NeurIPS 2021, in March 2022. \\
    
    \textcolor{\sectioncolor}{\textbf{
    What mechanisms or procedures were used to collect the data (e.g., hardware
    apparatus or sensor, manual human curation, software program, software
    API)?
    }
    How were these mechanisms or procedures validated?
    } \\
    The Argoverse 2 data comes from Argo `Z1' fleet vehicles. These vehicles use Velodyne lidars and traditional RGB cameras. All sensors are calibrated by Argo. HD maps and 3D object annotations are created and validated through a combination of computational tools and human annotations. Object tracks in the \emph{Motion Forecasting Dataset} are created by a 3D tracking algorithm.
    \\

    \textcolor{\sectioncolor}{\textbf{
    What was the resource cost of collecting the data?
    }
    (e.g. what were the required computational resources, and the associated
    financial costs, and energy consumption - estimate the carbon footprint.
    See Strubell \textit{et al.} for approaches in this area.)
    } \\
    The data was captured during normal fleet operations, so there was minimal overhead for logging particular events. The transformation and post-processing of several terabytes of data consumed an estimated 1,000 machine hours. We estimate a Carbon footprint of roughly 1,000 lbs based on the CPU-centric workload. \\
    
    \textcolor{\sectioncolor}{\textbf{
    If the dataset is a sample from a larger set, what was the sampling
    strategy (e.g., deterministic, probabilistic with specific sampling
    probabilities)?
    }
    } \\
    The \emph{Sensor Dataset} scenarios were chosen from a larger set through manual review. The \emph{Lidar Dataset} and \emph{Motion Forecasting Dataset} scenarios were chosen by heuristics which looked for interesting object behaviors during fleet operations. \\
    
    \textcolor{\sectioncolor}{\textbf{
    Who was involved in the data collection process (e.g., students,
    crowdworkers, contractors) and how were they compensated (e.g., how much
    were crowdworkers paid)?
    }
    } \\
    Argo employees and Argo interns curated the data. Data collection and data annotation was done by Argo employees. Crowdworkers were not used. \\
    
    \textcolor{\sectioncolor}{\textbf{
    Were any ethical review processes conducted (e.g., by an institutional
    review board)?
    }
    If so, please provide a description of these review processes, including
    the outcomes, as well as a link or other access point to any supporting
    documentation.
    } \\
    No. \\
    
    \textcolor{\sectioncolor}{\textbf{
    Does the dataset relate to people?
    }
    If not, you may skip the remainder of the questions in this section.
    } \\
    Yes. \\
    
    \textcolor{\sectioncolor}{\textbf{
    Did you collect the data from the individuals in question directly, or
    obtain it via third parties or other sources (e.g., websites)?
    }
    } \\
    The data is collected from vehicles on public roads, not from a third party.\\
    
    \textcolor{\sectioncolor}{\textbf{
    Were the individuals in question notified about the data collection?
    }
    If so, please describe (or show with screenshots or other information) how
    notice was provided, and provide a link or other access point to, or
    otherwise reproduce, the exact language of the notification itself.
    } \\
    No, but the data collection was not hidden. The Argo fleet vehicles are well marked and have obvious cameras and lidar sensors. The vehicles only capture data from public roads. \\
    
    \textcolor{\sectioncolor}{\textbf{
    Did the individuals in question consent to the collection and use of their
    data?
    }
    If so, please describe (or show with screenshots or other information) how
    consent was requested and provided, and provide a link or other access
    point to, or otherwise reproduce, the exact language to which the
    individuals consented.
    } \\
    No. People in the dataset were in public settings and their appearance has been anonymized. Drivers, pedestrians, and vulnerable road users are an intrinsic part of driving on public roads, so it is important that datasets contain people so that the community can develop more accurate perception systems. \\
    
    \textcolor{\sectioncolor}{\textbf{
    If consent was obtained, were the consenting individuals provided with a
    mechanism to revoke their consent in the future or for certain uses?
    }
     If so, please provide a description, as well as a link or other access
     point to the mechanism (if appropriate)
    } \\
    n/a \\
    
    \textcolor{\sectioncolor}{\textbf{
    Has an analysis of the potential impact of the dataset and its use on data
    subjects (e.g., a data protection impact analysis) been conducted?
    }
    If so, please provide a description of this analysis, including the
    outcomes, as well as a link or other access point to any supporting
    documentation.
    } \\
    No. \\
    
    \textcolor{\sectioncolor}{\textbf{
    Any other comments?
    }} \\
    n/a \\

\begin{mdframed}[linecolor=\sectioncolor]
\section*{\textcolor{\sectioncolor}{
    PREPROCESSING / CLEANING / LABELING
}}
\end{mdframed}

    \textcolor{\sectioncolor}{\textbf{
    Was any preprocessing/cleaning/labeling of the data
    done (e.g., discretization or bucketing, tokenization, part-of-speech
    tagging, SIFT feature extraction, removal of instances, processing of
    missing values)?
    }
    If so, please provide a description. If not, you may skip the remainder of
    the questions in this section.
    } \\
    Yes. Images are reduced from their full resolution. 3D point locations are quantized to float16. Ground height maps are quantized to .3 meter resolution from their full resolution. HD map polygon vertex locations are quantized to .01 meter resolution. 3D annotations are smoothed. For the \emph{Motion Forecasting Dataset}, transient 3D tracks are suppressed and object locations are smoothed over time. \\

    \textcolor{\sectioncolor}{\textbf{
    Was the “raw” data saved in addition to the preprocessed/cleaned/labeled
    data (e.g., to support unanticipated future uses)?
    }
    If so, please provide a link or other access point to the “raw” data.
    } \\
    Yes, but such data is not public. \\

    \textcolor{\sectioncolor}{\textbf{
    Is the software used to preprocess/clean/label the instances available?
    }
    If so, please provide a link or other access point.
    } \\
    No. \\

    \textcolor{\sectioncolor}{\textbf{
    Any other comments?
    }} \\
    n/a \\

\begin{mdframed}[linecolor=\sectioncolor]
\section*{\textcolor{\sectioncolor}{
    USES
}}
\end{mdframed}

    \textcolor{\sectioncolor}{\textbf{
    Has the dataset been used for any tasks already?
    }
    If so, please provide a description.
    } \\
    Yes, this manuscript benchmarks a contemporary 3D object detection method on the \emph{Sensor Dataset} and a contemporary motion forecasting method on the \emph{Motion Forecasting Dataset}. \\

    \textcolor{\sectioncolor}{\textbf{
    Is there a repository that links to any or all papers or systems that use the dataset?
    }
    If so, please provide a link or other access point.
    } \\
    Yes, the Argoverse 2 API can be found at \url{https://github.com/argoverse/av2-api}.

    For the Argoverse 2 datasets, we maintain two leaderboards for 3D Detection 
 [\url{https://eval.ai/web/challenges/challenge-page/1710}] and Motion Forecasting [\url{https://eval.ai/web/challenges/challenge-page/1719}].
    
    For the Argoverse 1 datasets, we maintain four leaderboards for 3D Tracking [\url{https://eval.ai/web/challenges/challenge-page/453/overview}], 3D Detection [\url{https://eval.ai/web/challenges/challenge-page/725/overview}], Motion Forecasting [\url{https://eval.ai/web/challenges/challenge-page/454/overview}], and Stereo Depth Estimation [\url{https://eval.ai/web/challenges/challenge-page/917/overview}]. Argoverse 1 was also used as the basis for a Streaming Perception challenge [\url{https://eval.ai/web/challenges/challenge-page/800/overview}]. \\

    \textcolor{\sectioncolor}{\textbf{
    What (other) tasks could the dataset be used for?
    }
    } \\
    The datasets could be used for research on visual odometry, pose estimation, lane detection, map automation, self-supervised learning, structure-from-motion, scene flow, optical flow, time to contact estimation, pseudo-lidar, and point cloud forecasting.\\

    \textcolor{\sectioncolor}{\textbf{
    Is there anything about the composition of the dataset or the way it was
    collected and preprocessed/cleaned/labeled that might impact future uses?
    }
    For example, is there anything that a future user might need to know to
    avoid uses that could result in unfair treatment of individuals or groups
    (e.g., stereotyping, quality of service issues) or other undesirable harms
    (e.g., financial harms, legal risks) If so, please provide a description.
    Is there anything a future user could do to mitigate these undesirable
    harms?
    } \\
    No. \\

    \textcolor{\sectioncolor}{\textbf{
    Are there tasks for which the dataset should not be used?
    }
    If so, please provide a description.
    } \\
    The dataset should not be used for tasks which depend on faithful appearance of faces or license plates since that data has been obfuscated. For example, running a face detector to try and estimate how often pedestrians use crosswalks will not result in meaningful data. \\

    \textcolor{\sectioncolor}{\textbf{
    Any other comments?
    }} \\
    n/a \\

\begin{mdframed}[linecolor=\sectioncolor]
\section*{\textcolor{\sectioncolor}{
    DISTRIBUTION
}}
\end{mdframed}

    \textcolor{\sectioncolor}{\textbf{
    Will the dataset be distributed to third parties outside of the entity
    (e.g., company, institution, organization) on behalf of which the dataset
    was created?
    }
    If so, please provide a description.
    } \\
    Yes, the dataset is hosted on \url{https://www.argoverse.org/} like Argoverse 1 and 1.1. \\

    \textcolor{\sectioncolor}{\textbf{
    How will the dataset will be distributed (e.g., tarball on website, API,
    GitHub)?
    }
    Does the dataset have a digital object identifier (DOI)?
    } \\
    We provide both tar.gz archives and raw files for two of the Argoverse 2 datasets (\emph{Motion Forecasting}, \emph{Sensor}), but provide only raw files for the \emph{Lidar} datasets), available via AWS transfer. See \url{https://www.argoverse.org/av2.html#download-link}. 

    The Argoverse 1 and Argoverse 1.1 were distributed as a series of tar.gz files (See \url{https://www.argoverse.org/av1.html#download-link}. The files are broken up to make the process more robust to interruption (e.g. a single 2 TB file failing after 3 days would be frustrating) and to allow easier file manipulation (an end user might not have 2 TB free on a single drive, and if they do they might not be able to decompress the entire file at once).\\

    \textcolor{\sectioncolor}{\textbf{
    When will the dataset be distributed?
    }
    } \\
    The data was made available for download after NeurIPS 2021, in March 2022. \\

    \textcolor{\sectioncolor}{\textbf{
    Will the dataset be distributed under a copyright or other intellectual
    property (IP) license, and/or under applicable terms of use (ToU)?
    }
    If so, please describe this license and/or ToU, and provide a link or other
    access point to, or otherwise reproduce, any relevant licensing terms or
    ToU, as well as any fees associated with these restrictions.
    } \\
    Yes, the dataset was released under the same Creative Commons license as Argoverse 1 -- CC BY-NC-SA 4.0. Details can be seen at \url{https://www.argoverse.org/about.html#terms-of-use}.\\

    \textcolor{\sectioncolor}{\textbf{
    Have any third parties imposed IP-based or other restrictions on the data
    associated with the instances?
    }
    If so, please describe these restrictions, and provide a link or other
    access point to, or otherwise reproduce, any relevant licensing terms, as
    well as any fees associated with these restrictions.
    } \\
    No. \\

    \textcolor{\sectioncolor}{\textbf{
    Do any export controls or other regulatory restrictions apply to the
    dataset or to individual instances?
    }
    If so, please describe these restrictions, and provide a link or other
    access point to, or otherwise reproduce, any supporting documentation.
    } \\
    No. \\

    \textcolor{\sectioncolor}{\textbf{
    Any other comments?
    }} \\
    n/a \\

\begin{mdframed}[linecolor=\sectioncolor]
\section*{\textcolor{\sectioncolor}{
    MAINTENANCE
}}
\end{mdframed}

    \textcolor{\sectioncolor}{\textbf{
    Who is supporting/hosting/maintaining the dataset?
    }
    } \\
    Argo AI \\

    \textcolor{\sectioncolor}{\textbf{
    How can the owner/curator/manager of the dataset be contacted (e.g., email
    address)?
    }
    } \\
    The Argoverse team responds through the Github page for the Argoverse 2 API: \url{https://github.com/argoverse/av2-api/issues}.
    
    The Argoverse team responds through the Github page for the Argoverse 1 API: \url{https://github.com/argoverse/argoverse-api/issues}. It currently contains 2 open issues and 126 closed issues.
    
    For privacy concerns, contact information can be found here: \url{https://www.argoverse.org/about.html#privacy} \\

    \textcolor{\sectioncolor}{\textbf{
    Is there an erratum?
    }
    If so, please provide a link or other access point.
    } \\
    No. \\

    \textcolor{\sectioncolor}{\textbf{
    Will the dataset be updated (e.g., to correct labeling errors, add new
    instances, delete instances)?
    }
    If so, please describe how often, by whom, and how updates will be
    communicated to users (e.g., mailing list, GitHub)?
    } \\
    It is possible that the constituent Argoverse 2 datasets are updated to correct errors. This was the case with Argoverse 1 which was incremented to Argoverse 1.1. Updates will be communicated on Github and through our mailing list. \\

    \textcolor{\sectioncolor}{\textbf{
    If the dataset relates to people, are there applicable limits on the
    retention of the data associated with the instances (e.g., were individuals
    in question told that their data would be retained for a fixed period of
    time and then deleted)?
    }
    If so, please describe these limits and explain how they will be enforced.
    } \\
    No. \\

    \textcolor{\sectioncolor}{\textbf{
    Will older versions of the dataset continue to be
    supported/hosted/maintained?
    }
    If so, please describe how. If not, please describe how its obsolescence
    will be communicated to users.
    } \\
    Yes. We still host Argoverse 1 even though we have declared it ``deprecated''. See \url{https://www.argoverse.org/av1.html#download-link}. We will use the same warning if we ever deprecate Argoverse 2. Note: Argoverse 2 does not deprecate Argoverse 1. They are independent collections of datasets. \\

    \textcolor{\sectioncolor}{\textbf{
    If others want to extend/augment/build on/contribute to the dataset, is
    there a mechanism for them to do so?
    }
    If so, please provide a description. Will these contributions be
    validated/verified? If so, please describe how. If not, why not? Is there a
    process for communicating/distributing these contributions to other users?
    If so, please provide a description.
    } \\
    Yes. For example, the streaming perception challenge was built by CMU researchers who added new 2D object annotations to Argoverse 1.1 data. The Creative Commons license we use for Argoverse 2 ensures that the community can do the same thing without needing Argo's permission.
    
    We do not have a mechanism for these contributions/additions to be incorporated back into the `base' Argoverse 2. Our preference would generally be to keep the `base' dataset as is, and to give credit to noteworthy additions by linking to them as we have done in the case of the Streaming Perception Challenge (see link at the top of this Argoverse page \url{https://www.argoverse.org/tasks.html}).\\

    \textcolor{\sectioncolor}{\textbf{
    Any other comments?
    }} \\
    n/a \\

\textbf{Environmental Impact Statement.} Amount of Compute Used: We estimate 2,000 CPU and 500 GPU hours were used in the collection of the dataset and the performance of baseline experiments.